\documentclass[runningheads]{llncs}

\usepackage{eccv}

\usepackage{eccvabbrv}

\usepackage{graphicx}
\usepackage{booktabs}

\usepackage{xcolor} 
\usepackage{paralist}
\usepackage[ruled,vlined]{algorithm2e}
\usepackage{multirow}
\usepackage{multicol}
\usepackage{array}
\usepackage{booktabs}
\usepackage[table]{xcolor}
\usepackage{amsmath}
\usepackage{makecell}
\usepackage{pifont}

\newcommand{\cmark}{\textcolor{green!50!black}{\ding{51}}} 
\newcommand{\xmark}{\textcolor{red!60!black}{\ding{55}}}   

\usepackage[accsupp]{axessibility}  

\usepackage{hyperref}

\usepackage{orcidlink}

\begin{document}

\title{{LogicIR}: Logic Gate Networks for Image Restoration}


\author{Hongjae Lee\orcidlink{0000-0001-5312-139X} \and
Myungjun Son\orcidlink{0009-0005-4577-4821} \and
Jaeseong Yu\orcidlink{0009-0005-1427-6651} \and
Seung-Won Jung\thanks{Corresponding author.}\orcidlink{0000-0002-0319-4467}}

\authorrunning{H. Lee \etal}

\institute{Korea University\\
\email{\{jimmy9704, sonbill, jsyu624, swjung83\}@korea.ac.kr}}

\maketitle

\begin{abstract}
Image restoration aims to reconstruct high-quality images from degraded low-quality inputs. As the computational demands of image restoration models continue to rise, there is growing interest in lightweight architectures optimized for fast and efficient inference. Logic gate networks (LGNs), which operate using fundamental logic operations such as NAND and XOR, have recently emerged as a promising direction for achieving highly efficient computation. However, their potential remains largely untapped in the domain of image restoration. In this work, we introduce LogicIR, the first LGN specifically designed for image restoration tasks. LogicIR incorporates a UNet-inspired architecture composed entirely of logic gates. In addition, we propose a differentiable bit decoding layer and an index shuffling mechanism that improves information propagation across logic gates. Experimental results across multiple image restoration benchmarks demonstrate that LogicIR achieves strong performance with significantly reduced computational cost, establishing LogicIR as a viable and efficient alternative for image restoration. The source code is available at our project page \url{https://github.com/jimmy9704/LogicIR}.
\end{abstract}

\section{Introduction}
\label{sec:intro}

\begin{figure}[t]
\centering
\begin{minipage}[t]{0.485\linewidth}
    \centering
    \includegraphics[width=\linewidth]{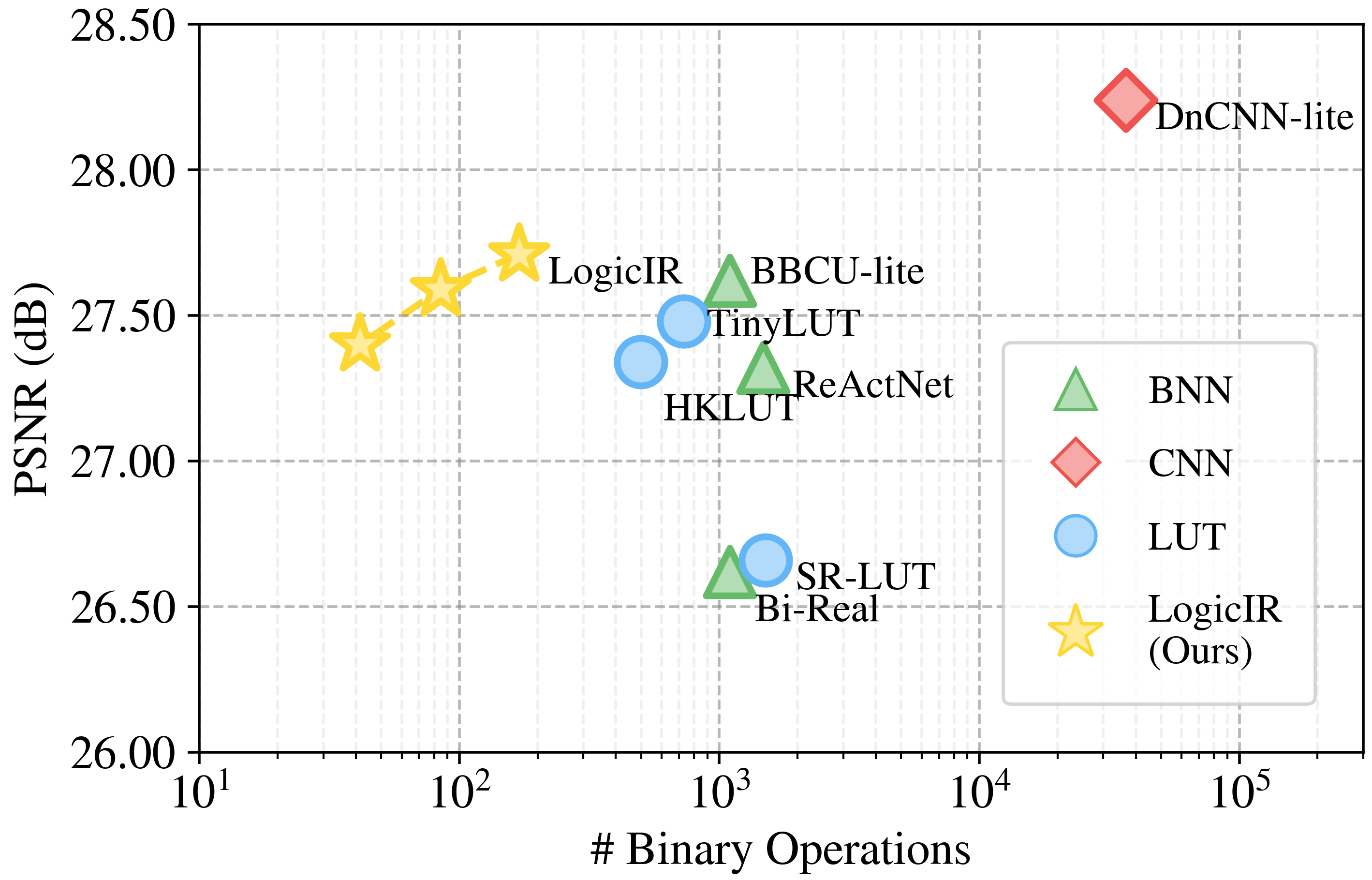}
    \caption{Trade-off between denoising performance and binary operation count on BSD68 ($\sigma=25$). Binary operation count is shown on a logarithmic scale.}
    \label{fig:fig1}
\end{minipage}\hfill
\begin{minipage}[t]{0.485\linewidth}
    \centering
    \includegraphics[width=\linewidth]{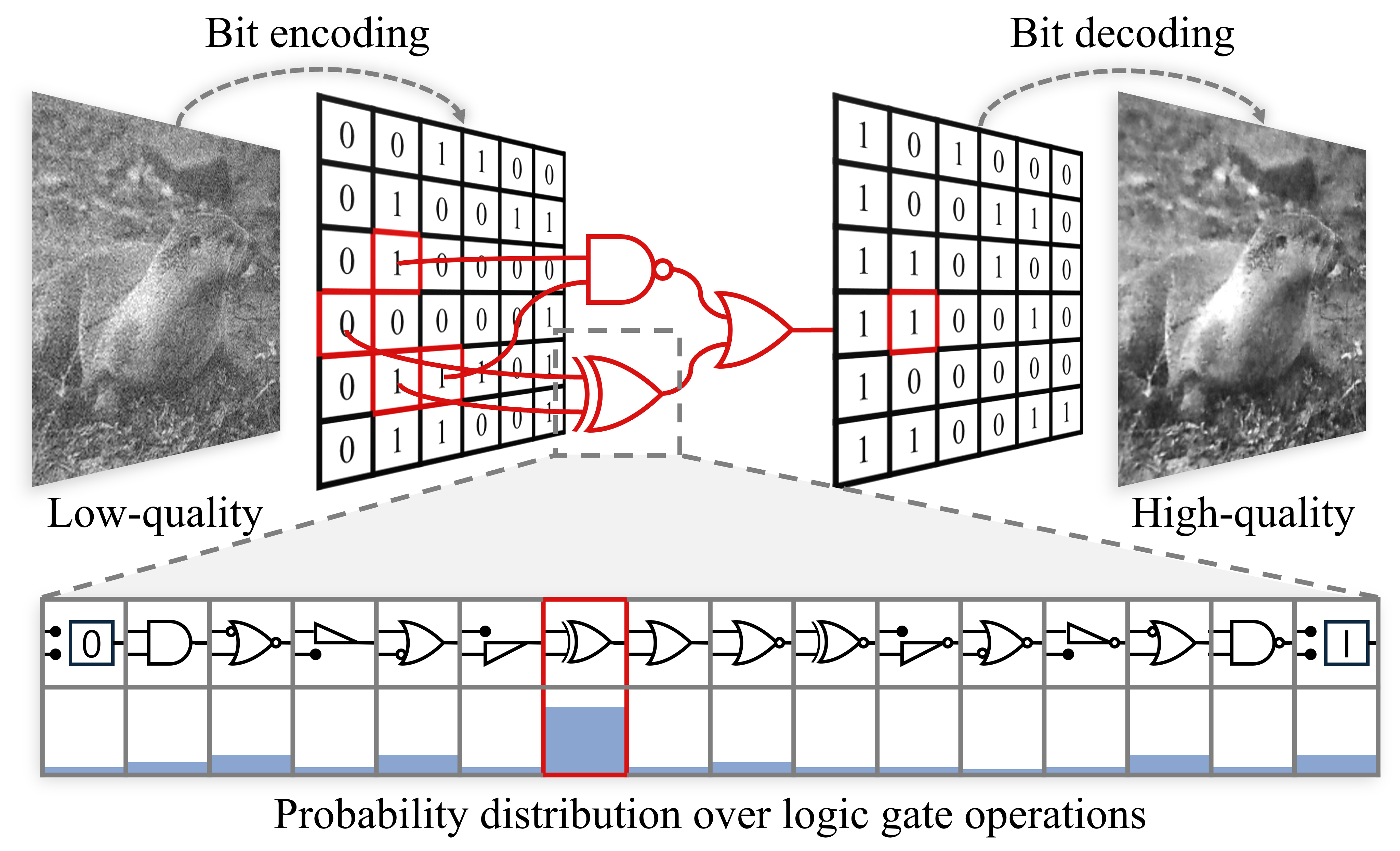}
    \caption{Overview of LogicIR using logic gates for image restoration. Each gate softly selects among 16 logic operations during training and applies the most probable one at inference.}
    \label{fig:overview}
  \centering
\end{minipage}
\end{figure}

Image restoration tasks, such as denoising, deblocking, and deblurring, aim to reconstruct high-quality images from degraded low-quality inputs. Although deep neural networks (DNNs)~\cite{zhang2017dncnn, swinir, zamir2022restormer} have led to significant improvements in image restoration, conventional DNNs typically incur substantial computational demands. Such high computational costs constrain their applicability in practical, resource-constrained environments, such as smartphones, wearable devices, and embedded systems. Consequently, extensive research has been conducted to develop computationally efficient image restoration models. Existing approaches include the use of lookup tables (LUTs)~\cite{jo2021srlut,mulut,hklut}, network pruning~\cite{zhang2022learning,shi2023memory}, quantization~\cite{refqsr,hong2024adabm}, and binary neural networks (BNNs)~\cite{bam,btm,bbcu,frbc}. 

In particular, BNNs utilize binary weights and activations, substantially reducing memory and computational demands by replacing multiplication operations with efficient binary operations. Recently, significant efforts have been directed toward improving BNNs for image restoration. For instance, BAM~\cite{bam} introduces a bit accumulation mechanism that iteratively refines binary weights and activations to approximate full-precision convolutions, while BBCU~\cite{bbcu} proposes a residual alignment strategy to mitigate the value-range mismatch between binary and full-precision features. In parallel, LUT-based methods~\cite{jo2021srlut,hklut,tinylut,im-lut} accelerate inference by training a compact network with a small receptive field and encoding its input–output mappings into a precomputed LUT. However, as shown in~\cref{fig:fig1}, BNN-based methods still rely on full-precision operations in several layers and involve a large number of trainable parameters, leading to high computational cost, whereas LUT-based approaches suffer from the exponential growth of LUT size, which restricts receptive field expansion and limits restoration performance.

To overcome these limitations, we explore logic gate networks (LGNs)~\cite{difflogic, ttnet, interp, convdiff}, which have recently emerged as a promising paradigm for efficient computation, and extend them to image restoration for the first time. LGNs perform inference exclusively using discrete logic operations, such as NAND and XOR, thereby eliminating the need for trainable weights. By operating directly at the logic-gate level, LGNs achieve exceptionally high computational efficiency and inference speed, especially when deployed on hardware platforms optimized for logic operations, such as FPGAs and ASICs. Furthermore, the explicit logical structure of LGNs improves the interpretability of their decision-making processes. However, LGNs were mainly developed for classification tasks, and thus lack the spatial modeling and hierarchical representation capabilities required for image restoration.

To this end, we propose LogicIR, the first LGN specifically designed for image restoration. As shown in~\cref{fig:fig1}, LogicIR significantly reduces computational complexity, measured by binary operation count, compared to existing lightweight approaches, while maintaining competitive performance. Unlike conventional methods that rely on arithmetic operations, LogicIR performs image restoration solely through logic operations, as depicted in~\cref{fig:overview}. To enable this, we design a UNet-inspired architecture composed entirely of logic gates, allowing hierarchical feature representation. Furthermore, we introduce a differentiable bit decoding layer that facilitates effective training of the logic gate-based image restoration models. Additionally, we introduce a novel index shuffling method to enhance information flow across logic gates. As a result, our approach achieves 27.71 dB in PSNR using only 169.3 G binary operations for denoising, demonstrating outstanding computational efficiency and competitive performance.

\section{Related Work}
\label{sec:related}

\textbf{Binarized neural networks for image restoration.} Recent studies have explored the binarization of neural networks to enable efficient image restoration~\cite{bam,btm,bbcu,frbc}. BTM~\cite{btm} proposed a BNN for super-resolution by removing batch normalization. Building upon this, BBCU~\cite{bbcu} extended the approach to broader restoration tasks, such as denoising and deblocking, by introducing a basic binary convolution unit. FRBC~\cite{frbc} enhanced the effectiveness of binarized networks through a distillation-guided training strategy that aligns the representations of the binary and full-precision models. However, these approaches still require storage for weight parameters and rely on full-precision adders within residual blocks, as well as non-binarized layers at the input and output, which hinder deployment on resource-constrained hardware platforms, such as FPGAs and ASICs.

\noindent \textbf{Look-up tables for image restoration.} Recent research has investigated the use of look-up table (LUT)-based methods to accelerate image restoration~\cite{jo2021srlut,mulut,li2024spflut,tinylut,hklut,im-lut}. SR-LUT~\cite{jo2021srlut} proposed a practical approach for image super-resolution by training a deep network with a small receptive field and encoding its input-output mappings into an LUT for efficient inference. MuLUT~\cite{mulut} extended this approach by leveraging multiple small LUTs to effectively increase the receptive field and generalized the method to various image restoration tasks. HKLUT~\cite{hklut} further reduced the LUT size by allocating different LUT resolutions based on the significance of bit planes in 8-bit input images. Despite these advancements, LUT-based methods suffer from a fundamental limitation: the size of the LUT grows exponentially with the receptive field. This imposes a practical constraint on the receptive field size, making it difficult for these methods to capture global spatial patterns and long-range dependencies.

\section{Background}
\textbf{Logic gate networks.} LGNs~\cite{difflogic, ttnet, interp, convdiff} are composed of binary logic gates such as AND, NAND, and XOR, operating at the lowest abstraction level of digital circuits. One of the key advantages of LGNs is their ability to perform inference using only simple logic operations, without relying on complex floating-point arithmetic, making them highly efficient for deployment on hardware platforms such as FPGAs and ASICs. Training LGNs involves a combinatorial optimization problem, as it requires selecting a logic gate for each node and defining the overall connection pattern. However, their non-differentiable nature precludes the use of gradient-based optimization methods. To overcome this limitation, a differentiable relaxation method has been recently introduced, enabling end-to-end training by learning probability distributions over logic gate selections~\cite{difflogic}. After training, the most probable gate is selected at each node, resulting in a discretized logic circuit well-suited for efficient hardware implementation.

\begin{figure}[t]
\centering
\begin{minipage}[t]{0.485\linewidth}
  \centering
  \captionof{table}{Logic gate network-based methods and other approaches on BSD68 ($\sigma=25$). BOPs: binary operations.}
  \scriptsize
  \setlength{\tabcolsep}{3.pt}
  \renewcommand{\arraystretch}{1.05}
    \begin{tabular}{lrcc}
        \toprule
        Method           & BOPs & RF size & PSNR \\
        \midrule
        DnCNN-lite~\cite{zhang2017dncnn}            & 36 562.6 G & 35 $\times$ 35 & 28.24 \\ 
        BBCU-lite~\cite{bbcu}        & 1 097.2 G & 35 $\times$ 35 & 27.62 \\ 
        HKLUT~\cite{hklut}    & 499.4 G & 5 $\times$ 5 & 27.34	\\
        \midrule
        StackedCLGN      &  \bf{39.7 G} & 25 $\times$ 25 & 17.19 \\ 
        \rowcolor{teal!10} LogicIR-S & 41.4 G & 47 $\times$ 47& 27.40  \\ 
        \rowcolor{teal!10} LogicIR-S-4RT & 169.3 G & 47 $\times$ 47& \bf{27.71}  \\ 
        \bottomrule
    \end{tabular}
    \label{tab:stack}
\end{minipage}\hfill
\begin{minipage}[t]{0.485\linewidth}
  \centering
  \vspace{.5\baselineskip}
  \includegraphics[width=\linewidth]{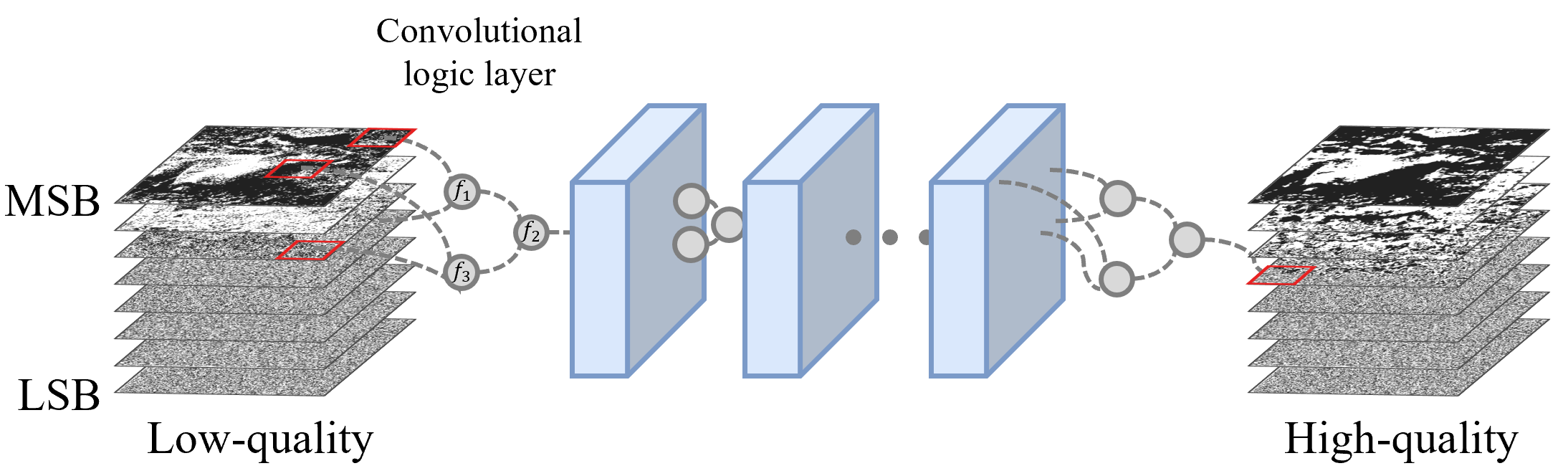}
    \caption{Overview of StackedCLGN, which stacks multiple convolutional logic layers. Each gray node represents a logic gate within a layer. The model processes 8-bit planes as input and output.}
    \label{fig:stack}
\end{minipage}
\end{figure}

\noindent \textbf{Convolutional logic gate networks.} 
Convolutional LGNs (CLGNs)~\cite{convdiff} address a critical limitation of conventional LGNs, which operate solely on flattened inputs and thus fail to capture spatial structure. To overcome this, CLGNs incorporate convolutional logic layers that employ hierarchical logic gate trees to extract local spatial correlations through localized logic operations. This design significantly enhances the network’s ability to exploit spatial patterns, leading to improvements in both accuracy and scalability for classification tasks. Each kernel in a convolutional logic layer is constructed as a complete binary tree of depth $d$, where each internal node performs a logic gate operation (\eg, NAND, XOR), and each leaf node receives inputs from a predefined receptive field of size $s_c\times s_k\times s_k$, with $s_c$ denoting the number of input channels and $s_k$ the kernel size. These connections are randomly initialized and remain fixed throughout training and inference.

For example, a logic tree kernel of depth $d=2$ requires three logic gates: two at the first layer and one at the second layer, as illustrated in~\cref{fig:overview}. Given four input binary activations, $a_1, a_2, a_3$, and $a_4$, sampled from a local receptive field, the output activation $\hat{a}$ is computed as:
\begin{equation}
    \hat{a} = f_3\big(\,f_1(a_1, a_2),\; f_2(a_3, a_4)\,\big),
\end{equation}
where each logic gate $f_i$ is selected from a softmax-based probability distribution over a predefined set of logic operations~\cite{difflogic, convdiff}. This formulation allows the CLGN to perform localized spatial reasoning similar to conventional convolutional neural networks (CNNs), while retaining the efficiency and interpretability of fully logic-based computation.

\noindent \textbf{Limitations.} 
While CLGNs can effectively extract local spatial patterns from images, they remain insufficient for image restoration tasks for several reasons. First, they lack a bit decoding mechanism required to reconstruct images from features extracted through logic operations. Second, simply stacking convolutional logic layers cannot effectively capture the hierarchical representations crucial for high-quality image restoration. Finally, image restoration tasks inherently require a strong locality inductive bias~\cite{2025cem, chen2023hat}, rendering randomly connected logic nodes with large receptive fields less effective. We address these limitations by introducing a UNet-based architecture, a differentiable bit decoding module, and an index shuffling mechanism, which together enable effective logic gate-based image restoration.

\section{Logic Gate Networks for Image Restoration}
\label{sec:method}

Image restoration aims to reconstruct a high-quality image from a degraded low-quality observation. Recent advances in CNNs have led to significant improvements in image restoration, primarily through the use of convolutional layers to extract complex feature representations. A straightforward way to apply LGNs to image restoration is to build a network by serially stacking multiple convolutional logic layers, a baseline we refer to as StackedCLGN, as shown in~\cref{fig:stack}. In this design, both the low-quality input and high-quality output images are converted into 8-channel binary representations corresponding to their bit planes. All convolutional logic kernels in StackedCLGN use a fixed tree depth of $d = 3$.

While StackedCLGN can extract local spatial features in a manner similar to CNNs and rely solely on logic operations for inference, it lacks key architectural components required for reconstructing high-quality images from degraded inputs. Specifically, it does not incorporate hierarchical feature representations, skip connections, and a dedicated bit decoding module for pixel value reconstruction, all of which are essential for effective image restoration. As shown in~\cref{tab:stack}, StackedCLGN significantly underperforms compared to CNN-based models such as DnCNN-lite~\cite{zhang2017dncnn}, and even BNN-based models like BBCU-lite~\cite{bbcu}, which still benefit from residual connections and hierarchical feature representations. These limitations motivate the development of LogicIR, an LGN architecture tailored for image restoration.

\begin{figure*}[!t]
  \centering
  \includegraphics[width=\linewidth]{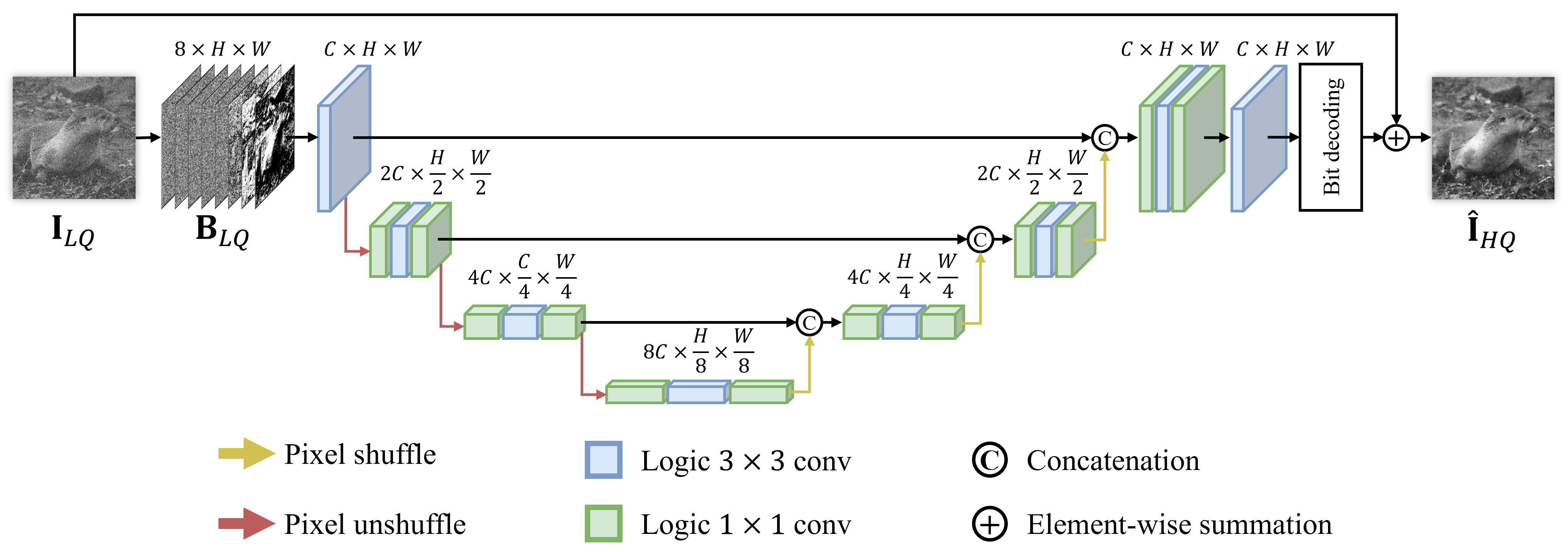}
  \caption{LogicIR architecture featuring a UNet-style encoder–decoder with convolutional logic layers. The model receives an 8-channel binary input and reconstructs the residual image using a differentiable bit decoding module.}
  \label{fig:arch}
\end{figure*}

\subsection{LogicIR Architecture} 

The overall architecture of LogicIR, illustrated in~\cref{fig:arch}, consists of two main components: a UNet-style~\cite{unet, mimo} encoder-decoder composed of convolutional logic layers and a bit decoder that reconstructs the residual image from bit predictions. Specifically, the low-quality input image  $\mathbf{I}_{LQ} \in \mathbb{R}^{H \times W}$ is first converted into an 8-channel binary representation, denoted as $ \mathbf{B}_{LQ} \in \{0,1\}^{8 \times H \times W} $. Each block in the encoder-decoder, except for the first and the last, comprises a $1 \times 1$ convolutional logic layer to capture inter-channel dependencies, followed by $3 \times 3$ and $1 \times 1$ convolutional logic layers to extract and refine spatial features. All convolutional logic layers throughout the network operate with a fixed tree depth of $d = 3$. Beyond simply replacing conventional convolutional layers with logic-based alternatives, LogicIR incorporates several architectural components specifically considered for LGN-based image restoration.

First, LogicIR avoids the use of addition operations commonly employed in conventional residual blocks~\cite{resnet,bireal}. To preserve high-frequency details and enable deeper training, it uses skip connections between the encoder and decoder stages via channel-wise concatenation. Second, for spatial resolution changes, LogicIR replaces conventional max pooling and transposed convolution operations with pixel unshuffle and pixel shuffle, respectively. These operations are not only well-suited for logic-based implementations, but also more effective in preserving spatial information. For example, pixel unshuffle transforms a feature map of shape ${C \times H \times W}$ into ${C r^2 \times H/r \times W/r}$, where the scale factor $r$ is set to 2. Finally, following the common practice in image restoration~\cite{zhang2017dncnn,rdn,mimo}, LogicIR is trained to predict a residual image $\mathbf{R} \in \mathbb{R}^{H \times W}$ rather than directly reconstructing the high-quality output image. The final output is then obtained by adding the predicted residual to the input low-quality image, i.e., $\hat{\mathbf{I}}_{HQ} = \mathbf{I}_{LQ} + \mathbf{R}$.

In addition to these design considerations, we introduce two novel components tailored to enhance image restoration with LGNs, which will be described in the following subsections.

\subsection{Bit Decoding} 
Although LogicIR can be configured to directly produce bit-planes representing pixel values, we found that this output format significantly complicates the training process. In particular, the less significant bit planes are often highly noisy and lack semantic consistency, which hinders the learning of meaningful gradients across all bits during pixel-value regression. To address this issue, we instead design LogicIR to produce a multi-channel binary output, which is subsequently transformed into a single-channel residual image using a bitcount operation defined as:\begin{equation}
    \mathbf{\bar{R}}(:,:) = \frac{\sum_{c=0}^{C-1} \mathbf{A}(c,:,:) - 0.5C}{0.5C},
\end{equation}
where $\mathbf{A} \in \{0,1\}^{C \times H \times W}$ denotes the final output activation map produced by the UNet decoder, and $\mathbf{\bar{R}} \in [-1, 1]^{H \times W}$ is the resulting normalized residual image. This formulation allows the logic-based network to learn in a more noise-robust manner. However, restricting the residual values to a fixed range may limit the network's representational capacity, potentially degrading performance. To mitigate this, we introduce a simple scaling mechanism that adapts the effective output range by obtaining the final residual image as $\mathbf{R} = \alpha \mathbf{\bar{R}}$, where $\alpha$ is a learnable scalar parameter.

\subsection{Index Shuffling} 
\begin{figure*}[!t]
  \centering
  \includegraphics[width=\linewidth]{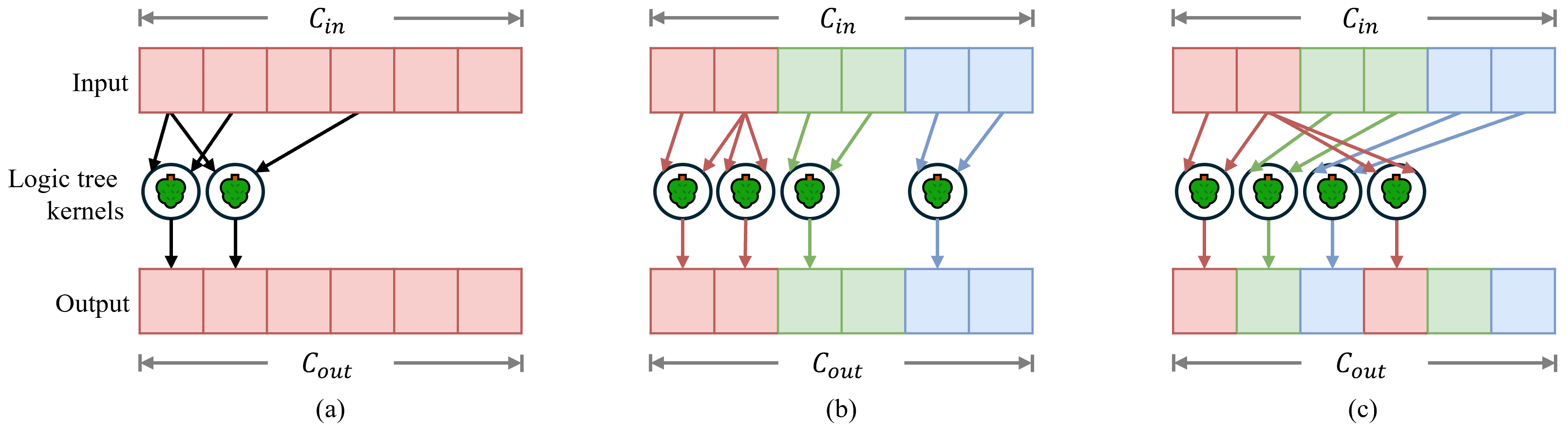}
  \caption{Index shuffling for convolutional logic layers with logic tree depth $d=1$. (a) Randomly connected without grouping. (b) Grouped logic layer with 3 groups, where each logic tree connects within a group. (c) Index shuffling cyclically reorders input indices.}
  \label{fig:index}
\end{figure*}

A convolutional logic layer is composed of multiple logic tree kernels. Each logic tree receives its input from randomly selected positions in the input activation map, as illustrated in~\cref{fig:index}(a). CLGN~\cite{convdiff}, inspired by the grouped convolution~\cite{alexnet}, introduces a constraint on these random connections by limiting them within fixed groups (\cref{fig:index}(b)). While this strategy enables a form of structured randomness, it also presents a limitation: group isolation restricts inter-group information flow, effectively forming independent sub-networks.

To address this issue, we draw inspiration from ShuffleNet~\cite{zhang2018shufflenet} and introduce an index shuffling mechanism in LGNs. As shown in~\cref{fig:index}(c), our method achieves inter-group information exchange by reordering the input indices connected to each logic tree kernel. Specifically, let $\mathbf{A}_{in} \in \{0,1\}^{C_{in} \times H \times W}$ denote an input activation map with $C_{in}$ channels. The input channels are divided into $G$ groups, where $G = C_{in} / 2^d$ and $d$ denotes the depth of the logic tree kernel. This grouping can be expressed as:
\begin{equation}
    \mathbf{A}_{in} = \{\mathbf{A}_{in}^0 \mid \mathbf{A}_{in}^1 \mid \cdots \mid \mathbf{A}_{in}^{G-1}\}, \quad \mathbf{A}_{in}^g \in \{0,1\}^{2^d \times H \times W},
\end{equation}
where $\mid$ represents channel-wise concatenation. Let $\mathbf{A}_{out} \in \{0,1\}^{C_{out} \times H \times W}$ denote an output activation map with $C_{out}$ channels. Each output channel is computed by applying a logic tree kernel to a cyclically selected input group, as follows:
\begin{equation}
    \mathbf{A}_{out}(n,:,:) = \mathcal{F}^{(n)}(\mathbf{A}_{in}^{g_{cyc}}), \quad \text{for } n = 0, 1, \dots, C_{out}-1
\end{equation}
where $g_{cyc} = n \bmod G$, and $\mathcal{F}^{(n)}$ represents the logic tree kernel for the $n$-th output channel. 

\subsection{Training} 
\textbf{Loss functions.} Following prior works in image restoration~\cite{zhang2017dncnn, mulut}, we adopt the L2 loss, denoted as $\mathcal{L}_\text{2}$, as the training objective, which is defined as the pixel-wise difference between the reconstructed output and the ground-truth image: 
\begin{equation}
    \mathcal{L}_\text{2} = \left\| {\mathbf{I}}_{HQ} - \hat{\mathbf{I}}_{HQ} \right\|_2~,
\end{equation}
where $\hat{\mathbf{I}}_{HQ}$ and $\mathbf{I}_{HQ}$ denote the predicted and ground-truth images, respectively, and $\left\| \cdot \right\|_2$ represents the L2 norm. This loss formulation encourages accurate reconstruction by penalizing large deviations in pixel intensity.
 
In addition to the primary reconstruction loss $\mathcal{L}_\text{2}$, we incorporate an auxiliary supervision that accounts for the varying significance of individual bit planes. Specifically, we reconstruct a reference image $\tilde{\mathbf{I}}_{HQ}$ using only the top four most significant bit (MSB) planes of the ground-truth image $\mathbf{I}_{HQ}$ and measure the MSB loss, denoted as $\mathcal{L}_{\text{MSB}}$, as follows: 
\begin{equation}
    \mathcal{L}_{\text{MSB}} =  \left\| \tilde{\mathbf{I}}_{HQ} - \hat{\mathbf{I}}_{HQ} \right\|_2.
\end{equation}
This auxiliary term encourages the model to prioritize information contained in the most informative bit planes, which are critical for preserving structural details. The final training objective is defined as a weighted sum of the two loss terms:
\begin{equation}
    \mathcal{L}_{\text{total}} = \mathcal{L}_{2} + \lambda \mathcal{L}_{\text{MSB}},
\end{equation}
where $\lambda$ is a hyperparameter that controls the relative contribution of the MSB loss.

\noindent \textbf{Fine-tuning using straight-through estimators.} During training, each logic gate in the differentiable LGNs~\cite{difflogic, convdiff} performs a soft selection over all 16 possible binary logic operations, with each operation weighted by its learned probability. In contrast, during inference, the gate applies a discrete selection, choosing the operation with the highest probability. The computation of the logic gate in both training and inference phases can be expressed as:
\begin{equation}
    \hat{a}= 
    \begin{cases}
        \sum_{i=0}^{15} p_i \cdot f_i(a_1, a_2), & \text{Training}, \\
        f_{i^*}(a_1, a_2), \quad \text{where } i^* = \displaystyle\arg\max_{i} p_i, & \text{Inference}.
    \end{cases}
\end{equation}
Here, $f_i$ denotes the $i$-th logic operation, and $p_i$ is its corresponding learned probability. The soft formulation during training allows gradients to flow through all candidate logic operations.

However, this mismatch between soft logic selection during training and hard logic selection during inference can lead to a performance gap. To address this issue, we introduce an additional fine-tuning strategy based on the straight-through estimator (STE)~\cite{ste}, applied after the main training phase. Specifically, in the fine-tuning stage, hard logic operations are used in the forward pass, while gradients are computed as if the soft combination were still applied. This enables the network to better adapt to the discrete behavior used at inference time. In addition, the scaling parameter $\alpha$, which modulates the range of the residual output, is also optimized during this fine-tuning stage.

\noindent \textbf{Rotational ensemble.} Due to the random connectivity in the logic gate network, each logic tree is exposed to a limited set of patterns within its receptive field. To address this limitation, we adopt a rotational ensemble strategy, in which the input image is rotated at multiple fixed angles, and the corresponding outputs are averaged to produce the final result. This approach increases the diversity of patterns seen by the network without modifying its structure. Formally, let $\mathcal{R}_\theta(\cdot)$ denote a rotation operation by angle $\theta$, and let $\mathcal{G}(\cdot)$ be the LogicIR model. The final output $\hat{\mathbf{I}}$ is computed as:

\begin{equation}
    \hat{\mathbf{I}}_{HQ} = \frac{1}{|\Theta|} \sum_{\theta \in \Theta} \mathcal{R}_{-\theta} \big( \mathcal{G}( \mathcal{R}_\theta(\mathbf{I}_{LQ}) ) \big),
\end{equation}

where $\Theta$ is the set of rotation angles, and $\mathcal{R}_{-\theta}(\cdot)$ denotes the inverse rotation used to restore the original orientation. To enable this ensemble strategy, we perform additional rotation-aware training, initialized from a pre-trained model originally trained without rotation.

\section{Experiments} 

\subsection{Experiment Settings}
\textbf{Implementation details.}
Following DnCNN~\cite{zhang2017dncnn}, we use 400 grayscale images from the BSD dataset~\cite{schmidt2014shrinkage} for training. The deraining models are trained on the Rain13K dataset~\cite{rain13k}. When processing color images, LogicIR treats the R, G, and B channels independently and processes them in parallel. All models are trained for $8 \times 10^4$ iterations using the Adam optimizer with a learning rate of $10^{-2}$. For tasks other than denoising, such as JPEG deblocking and deraining, we initialize the network using the model trained on the denoising task.

\noindent \textbf{Evaluation settings.}
Following BBCU~\cite{bbcu}, BBCU-lite and other BNN-based methods use DnCNN-lite as their baseline, a lightweight variant of DnCNN with 12 channels. To evaluate both image restoration quality and hardware efficiency, we adopt PSNR, CLIP-IQA~\cite{clipiqa}, LPIPS~\cite{lpips}, NIQE~\cite{niqe}, and SSIM as image quality metrics. We report the binary operation count (BOPs)~\cite{difflogic,ttnet,convdiff}, measured on $1280 \times 720$ input images. Similar to FLOPs, it represents the hardware cost and resource usage in image processing. Details on BOPs counting are provided in prior work~\cite{difflogic,ttnet,convdiff} and the supplementary material. In addition, Power, Area, and Runtime are measured to assess hardware efficiency.

\begin{table}[!t]
  \caption{Quantitative results for image denoising with a noise level of $\sigma = 25$ and JPEG deblocking with a quality factor of $ q = 10$. BOPs (binary operation count) are measured on $1280 \times 720$ input images. BBCU-lite-fully denotes a network with all layers binarized. LogicIR-S-2RT and LogicIR-S-4RT denote models applying rotational ensembles with 2 and 4 rotations, respectively.}
  \label{tab:main}
  \centering
  \scriptsize
  \setlength{\tabcolsep}{3.pt}
  \renewcommand{\arraystretch}{1.05}
  \begin{tabular}{c l rccc c rcc}
    \toprule
    \multirow{3}{*}{} &\multirow{3}{*}{Method} & \multicolumn{4}{c}{Denoising} && \multicolumn{3}{c}{Deblocking} \\
    \cmidrule(lr){3-6}
    \cmidrule(lr){8-10}
    & &  \multicolumn{1}{c}{\multirow{2}{*}{BOPs}} & \multicolumn{3}{c}{PSNR} && \multicolumn{1}{c}{\multirow{2}{*}{BOPs}} & \multicolumn{2}{c}{PSNR} \\
    & & & BSD68 & Set12 & Urban100 & && LIVE1 & Classic5 \\
    \midrule
    \multirow{3}{*}{\rotatebox{90}{FP}}
    &DnCNN-lite~\cite{zhang2017dncnn}       & 36.6 T    & 28.24 & 29.05 & 27.73    && 36.6 T    & 28.78 & 28.89 \\
    &DnCNN~\cite{zhang2017dncnn}            & 1 023.1 T & 29.23 & 30.44 & 29.95    && 1 023.1 T & 29.40 & 29.19 \\
    &SwinIR~\cite{swinir}           & 21 151.7 T & 29.50 & 31.01 & 31.30    && 21 151.7 T & 29.86 & 30.27 \\
    \midrule

    \multirow{5}{*}{\rotatebox{90}{BNN}}
    &BBCU-lite-fully~\cite{bbcu}  & 700.6 G       & 25.23 & 25.39 & 24.81    && 700.6 G        & 28.13 & 28.27 \\
    &Bi-Real~\cite{bireal}          & 1 095.3 G     & 26.62 & 26.96 & 26.11    && 1 095.3 G      & 28.31 & 28.39 \\
    &BBCU-lite~\cite{bbcu}        & 1 097.2 G     & 27.62 & 28.22 & 26.84 && 1 097.2 G      & 28.43 & 28.57 \\
    &ReActNet~\cite{liu2020reactnet}         & 1 471.4 G     & 27.32 & 27.91 & 26.53    && 1 471.4 G      & 28.40 & 28.51 \\
    &BBCU~\cite{bbcu}             & 9 152.1 G    & 28.45 & 29.33 & 28.16 && 9 152.1 G     & 29.06 & 30.00 \\
    \midrule

    \multirow{3}{*}{\rotatebox{90}{LUT}}
    &HKLUT~\cite{hklut}            & 499.4 G       & 27.34 & 27.94 & 26.30 && 502.7 G        & 28.54 & 28.65 \\
    &TinyLUT~\cite{tinylut}            & 729.8 G &27.48 & 28.26  & 26.42  &&  715.4 G &28.52 & 28.63 \\
    &SR-LUT~\cite{jo2021srlut}           & 1 505.5 G     & 26.66 & 27.10 & 25.94    && 1 693.6 G   & 28.51 & 28.61 \\
    \midrule

    \multirow{4}{*}{\rotatebox{90}{Logic}}
    & \cellcolor{teal!10} LogicIR-S
    & \cellcolor{teal!10}\bf{41.4 G}
    & \cellcolor{teal!10}27.40
    & \cellcolor{teal!10}27.83
    & \cellcolor{teal!10}26.57
    &\cellcolor{teal!10}& \cellcolor{teal!10}\bf{45.2 G}
    & \cellcolor{teal!10}28.48
    & \cellcolor{teal!10}28.52 \\
    & \cellcolor{teal!10} LogicIR-L
    & \cellcolor{teal!10}78.6 G
    & \cellcolor{teal!10}27.50
    & \cellcolor{teal!10}27.94
    & \cellcolor{teal!10}26.69
    &\cellcolor{teal!10}& \cellcolor{teal!10}73.6 G
    & \cellcolor{teal!10}28.54
    & \cellcolor{teal!10}28.57 \\
    & \cellcolor{teal!10} LogicIR-S-2RT
    & \cellcolor{teal!10}84.7 G
    & \cellcolor{teal!10}27.59
    & \cellcolor{teal!10}28.07
    & \cellcolor{teal!10}26.75
    &\cellcolor{teal!10}& \cellcolor{teal!10}92.3 G
    & \cellcolor{teal!10}28.57
    & \cellcolor{teal!10}28.61 \\
    & \cellcolor{teal!10} LogicIR-S-4RT
    & \cellcolor{teal!10}169.3 G
    & \cellcolor{teal!10}\bf{27.71}
    & \cellcolor{teal!10}\bf{28.22}
    & \cellcolor{teal!10}\bf{26.85}
    &\cellcolor{teal!10}& \cellcolor{teal!10}184.6 G
    & \cellcolor{teal!10}\bf{28.62}
    & \cellcolor{teal!10}\bf{28.66} \\
    \bottomrule
  \end{tabular}
\end{table}

\begin{figure}[!t]
    \centering
    \includegraphics[width=\linewidth]{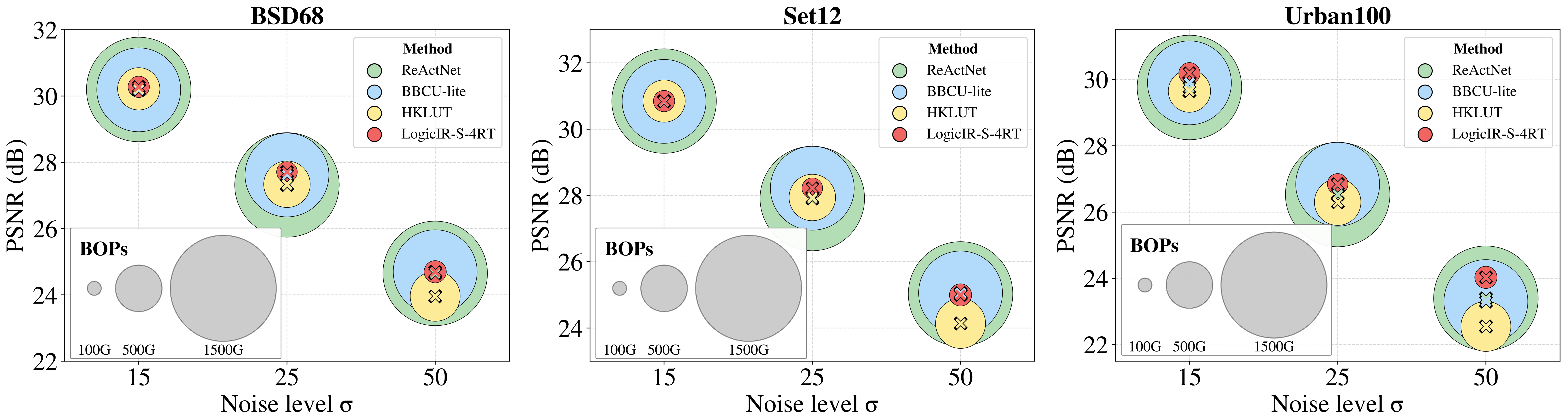}
    \caption{Denoising performance versus complexity (BOPs) across noise levels on BSD68, Set12, and Urban100. Bubble area is proportional to BOPs, and the $\times$ marker denotes the center of each bubble.}
    \label{fig:noise_level}
\end{figure}

\subsection{Evaluation on Image Denoising} 

We compare LogicIR with various baseline methods, including full-precision models (DnCNN, SwinIR), BNNs (BBCU, Bi-Real, ReActNet), and LUT-based approaches (HKLUT, TinyLUT, SR-LUT). LogicIR-S and LogicIR-L refer to the small and large variants of our model, using $C$ = 2048 and $C$ = 4096 channels in~\cref{fig:arch}, respectively.

\noindent \textbf{Quantitative results.} \cref{tab:main} presents quantitative grayscale image denoising results. LogicIR-S achieves comparable PSNR to ReActNet with only 2.8\% of its BOPs, while LogicIR-L outperforms the LUT-based HKLUT using just 15.7\% of its BOPs. Rotational ensembles further improve performance: LogicIR-S-4RT reaches 27.71~dB on BSD68 with 169.3~G BOPs, surpassing BBCU-lite (27.62~dB) with 6.5$\times$ fewer operations. Unlike prior BNNs (e.g., Bi-Real, ReActNet, BBCU-lite) that keep the first and last layers in full precision, LogicIR is fully binarized, making it natively compatible with logic-based implementations; in contrast, fully binarized variants of those BNNs (e.g., BBCU-lite-fully) incur large performance drops. As shown in~\cref{fig:noise_level}, across all three noise levels ($\sigma=15, 25, 50$), LogicIR consistently achieves comparable PSNR while using the fewest BOPs among all methods.

\begin{figure}[!t]
    \centering
    \includegraphics[width=\linewidth]{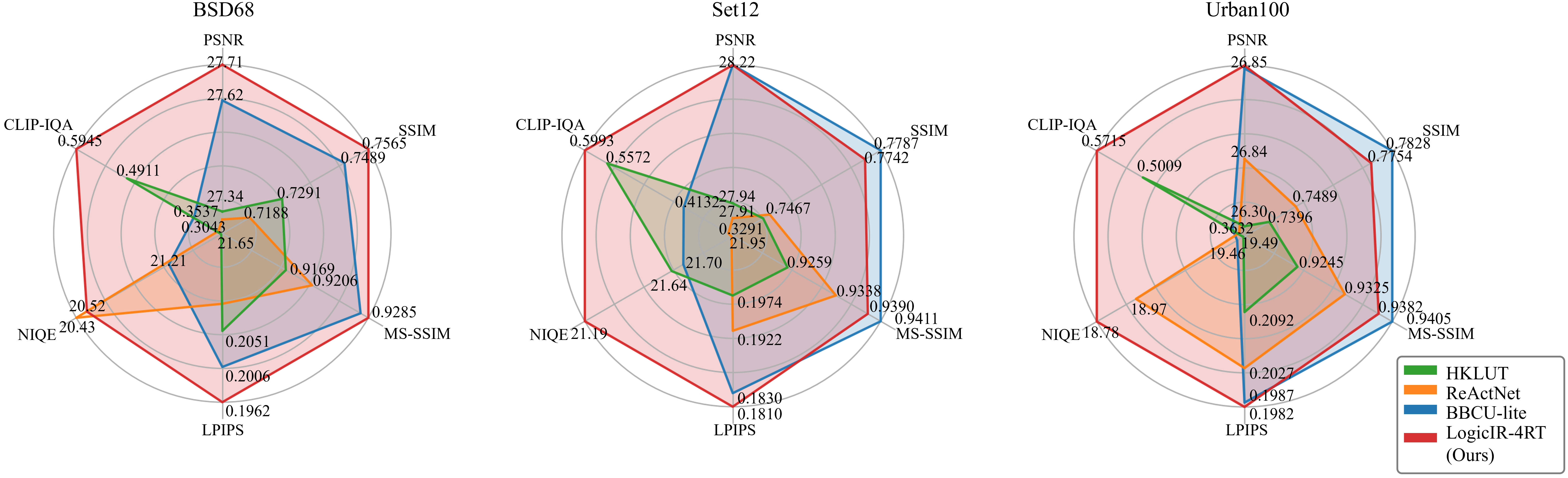}
    \caption{Comparison of image quality metrics for various methods with a noise level of $\sigma = 25$.}
    \label{fig:quality}
\end{figure}

\begin{figure*}[!t]
  \centering

  \includegraphics[width=\linewidth]{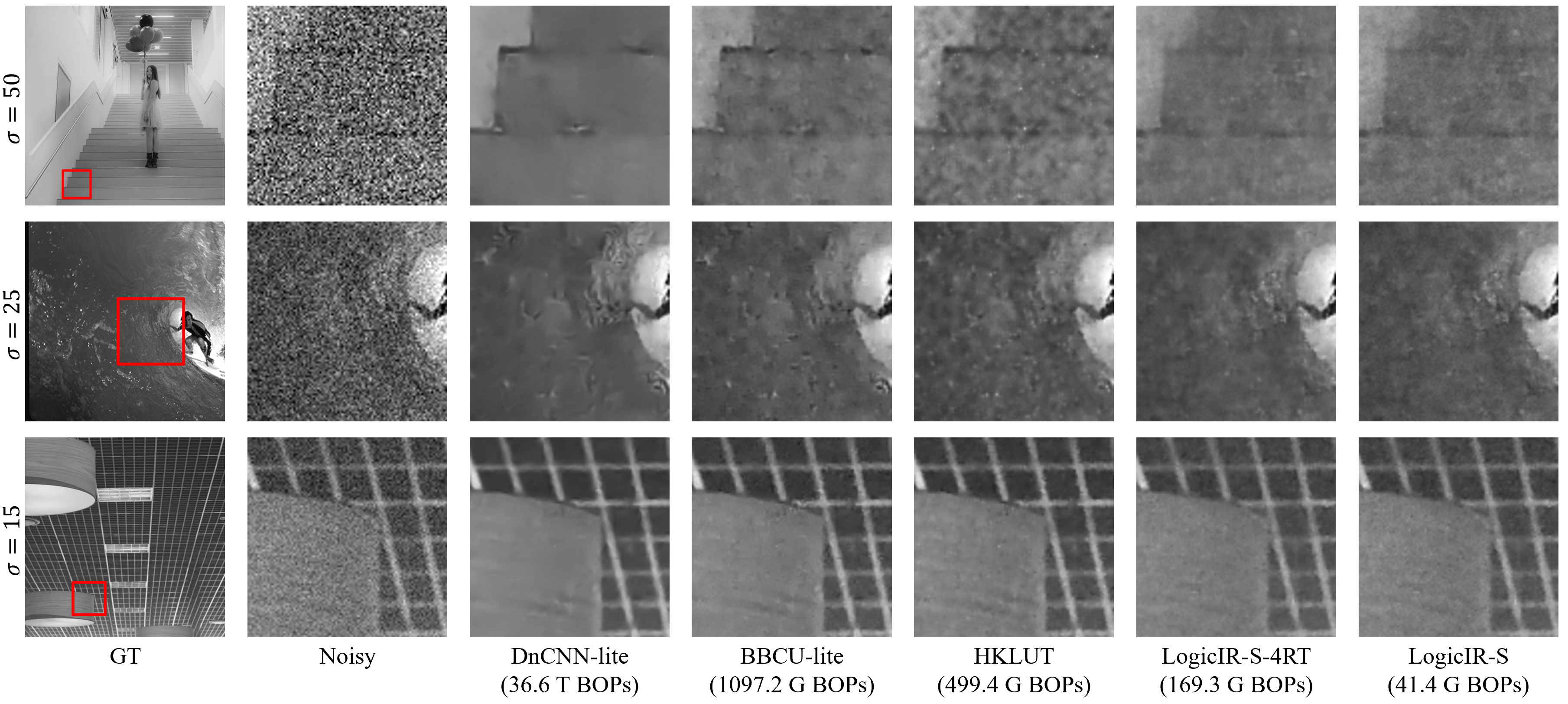}
  \captionof{figure}{Qualitative results on image denoising.}
  \label{fig:denoising}

  \vspace{0.5em}

  \includegraphics[width=\linewidth]{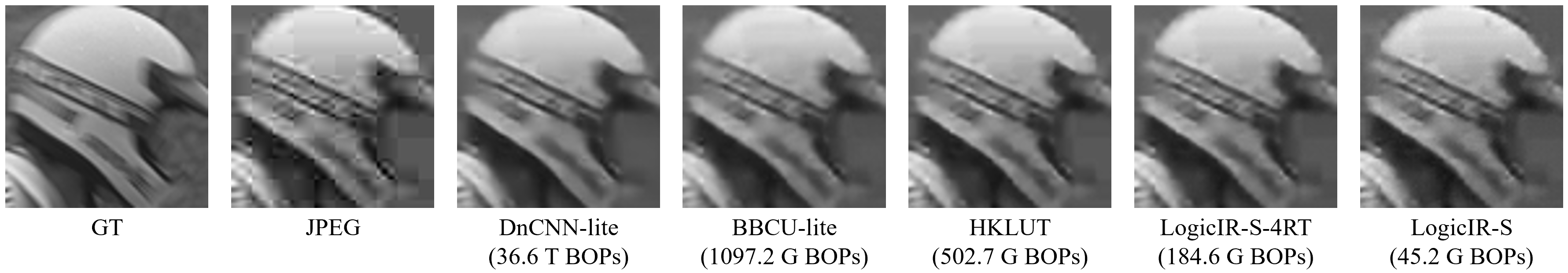}
  \captionof{figure}{Qualitative results on JPEG deblocking.}
  \label{fig:deblocking}

  \vspace{0.5em}

  \includegraphics[width=\linewidth]{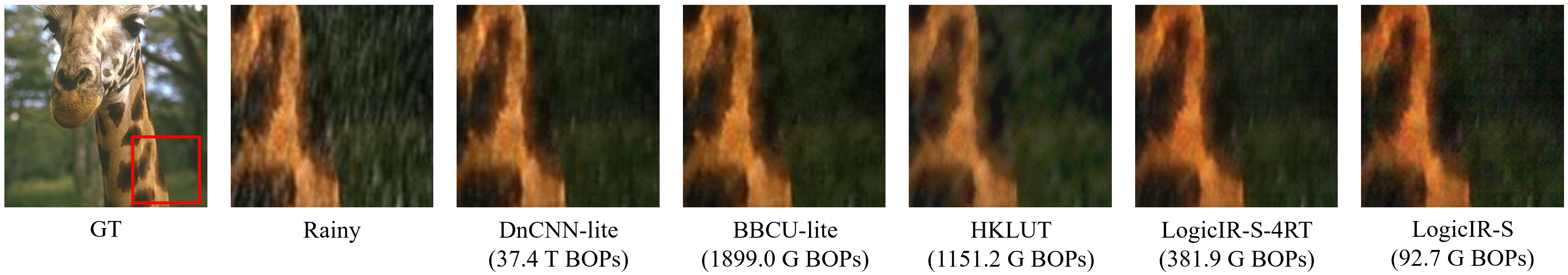}
  \captionof{figure}{Qualitative results on image deraining.}
  \label{fig:deraining}

\end{figure*}

\noindent \textbf{Qualitative and perceptual evaluation.} \cref{fig:quality} compares LogicIR with other methods using distortion metrics (PSNR, SSIM, MS-SSIM) and perceptual metrics (LPIPS, NIQE, CLIP-IQA). LogicIR-S-4RT achieves superior overall quality with far fewer binary operations. LUT-based methods (e.g., HKLUT) show weaker perceptual quality due to limited receptive fields, whereas LogicIR’s hierarchical logic design effectively restores fine details with high efficiency. Qualitative examples in~\cref{fig:denoising} further show that LogicIR yields visually comparable results to lightweight models such as BBCU-lite, DnCNN-lite, and HKLUT despite using fewer operations. Additional qualitative and quantitative results are provided in the supplementary material.

\subsection{Evaluation on Image Deblocking}
We further evaluate deblocking performance at a quality factor of $q = 10$, using the LIVE1~\cite{live1} and Classic5~\cite{classic5} datasets, as presented in~\cref{tab:main}. LogicIR-S achieves competitive performance, outperforming existing BNN-based models such as ReActNet and BBCU-lite. The larger variant, LogicIR-S-4RT, incorporates four rotations, achieves a PSNR of 28.62 dB on LIVE1, outperforming the LUT-based HKLUT (28.54 dB) while using only 36.7\% of its operations. These results highlight the high efficiency of LogicIR in the deblocking task. As shown in~\cref{fig:deblocking}, LogicIR effectively reduces compression artifacts and preserves structural details, delivering visual quality on par with other methods while operating with significantly fewer operations.

\subsection{Evaluation on Image Deraining}
We further evaluate the performance of LogicIR on the image deraining task on the Test100~\cite{test100} dataset, as presented in~\cref{tab:derain}. LogicIR-S achieves competitive results compared to the existing LUT-based HKLUT. For instance, LogicIR-S achieves a PSNR of 22.75 dB, comparable to HKLUT (22.71 dB), while using only 8.1\% of its operations. Considering an even larger variant of our model, LogicIR-S-4RT achieves a PSNR of 22.95 dB, surpassing all BNN-based methods while maintaining high computational efficiency, using only 16.8\% to 20.2\% of their operations. Qualitative deraining comparisons are shown in~\cref{fig:deraining}. We additionally provide results on another restoration task, low-light image enhancement, in the supplementary material.

\begin{table}[t]
\centering
\begin{minipage}[t]{0.445\columnwidth}
\caption{Quantitative comparison for image deraining.}
\centering
\scriptsize
\setlength{\tabcolsep}{1.5pt}
\renewcommand{\arraystretch}{.93}
\begin{tabular}{l rcc}
    \toprule
    Method & BOPs & PSNR & SSIM \\
    \midrule
    DnCNN-lite~\cite{zhang2017dncnn}    & 37.4 T & 23.34 & 0.7828 \\
    PreNet~\cite{prenet}        & 15 025.3 T & 24.81 & 0.8510 \\
    \midrule
    Bi-Real~\cite{bireal}        & 1 893.4 G & 22.79 & 0.7410 \\
    BBCU-lite\cite{bbcu}      & 1 899.0 G & 22.81 & 0.7647 \\
    ReActNet~\cite{liu2020reactnet}       & 2 269.5 G & 22.77 & 0.7562 \\
    \midrule
    HKLUT~\cite{hklut}          & 1 151.2 G & 22.71 & 0.7422 \\
    \midrule
    \rowcolor{teal!10} LogicIR-S      & \bf{92.7 G} & 22.75 & 0.7403 \\
    \rowcolor{teal!10} LogicIR-S-4RT  & 381.9 G & \bf{22.95} & \bf{0.7651} \\
    \bottomrule
\end{tabular}
\label{tab:derain}
\end{minipage}
\hfill
\begin{minipage}[t]{0.525\columnwidth}
\caption{Ablation study of LogicIR components on image denoising.}
\centering
\scriptsize
\setlength{\tabcolsep}{3pt}
\renewcommand{\arraystretch}{.9}
\begin{tabular}{lr cccccc}
\toprule
    Method                                & PSNR & \rotatebox{90}{bit}~\rotatebox{90}{decoding} & ~~\rotatebox{90}{UNet}~~ & \rotatebox{90}{index}~\rotatebox{90}{shuffling} & \rotatebox{90}{MSB loss} & \rotatebox{90}{fine-tuning}   \\
\midrule
    StackedCLGN    &  17.19& \xmark & \xmark & \xmark & \xmark & \xmark \\
    + Bit decoding &  26.83 & \cmark & \xmark & \xmark & \xmark & \xmark \\
    + UNet backbone & 27.15 & \cmark & \cmark & \xmark & \xmark & \xmark \\
    + Index shuffling & 27.58 & \cmark & \cmark & \cmark & \xmark & \xmark \\
    + MSB loss        & 27.72 & \cmark & \cmark & \cmark & \cmark & \xmark \\
\rowcolor{teal!10}
    LogicIR-S & \bf{27.83} & \cmark & \cmark & \cmark & \cmark & \cmark \\
\bottomrule
\end{tabular}
\label{tab:ablation}
\end{minipage}
\end{table}

\begin{figure}[t]
\centering
\begin{minipage}[t]{0.445\linewidth}
  \centering
  {\captionsetup{type=table, belowskip=10pt}%
  \captionof{table}{Ablation of color-processing strategies in LogicIR. Y-only uses Y in YCbCr. RGB-joint restores RGB jointly with a widened network. RGB-channel uses per-channel R/G/B with shared weights.}
  \label{tab:ablation_color}
  }
  \scriptsize
  \setlength{\tabcolsep}{4.pt}
  \renewcommand{\arraystretch}{1.4}
    \begin{tabular}{l rcc}
        \toprule
        Method & BOPs & PSNR & SSIM \\
        \midrule
        Y-only       &  \bf{35.9 G} & 22.63 & 0.7394 \\
        RGB-joint    & 169.1 G & \bf{22.79} & \bf{0.7422}     \\
        RGB-channel     & 92.7 G & 22.75 & 0.7403\\
        \bottomrule
    \end{tabular}
\end{minipage}\hfill
\begin{minipage}[t]{0.525\linewidth}
  \centering
  \vspace{.5\baselineskip}
  \includegraphics[width=\linewidth]{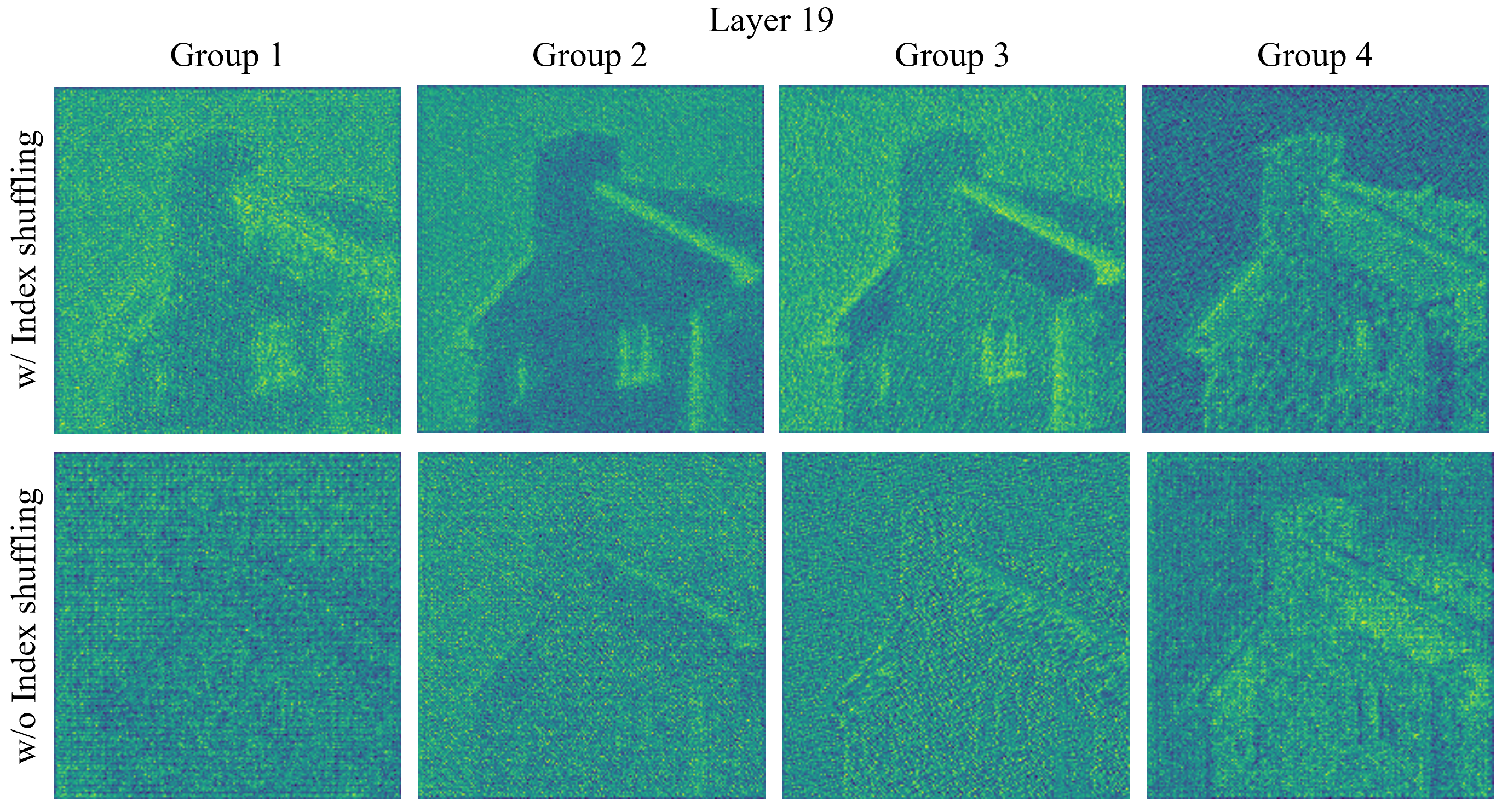}
    \caption{Group-wise averaged feature maps at layer 19 with and without index shuffling, showing enhanced representations.}
    \label{fig:ablation}
\end{minipage}
\end{figure}

\subsection{Ablation Studies} 

\begin{figure}[!t]
  \centering
  \begin{minipage}[t]{0.45\linewidth}
    \centering
    \includegraphics[width=.85\linewidth]{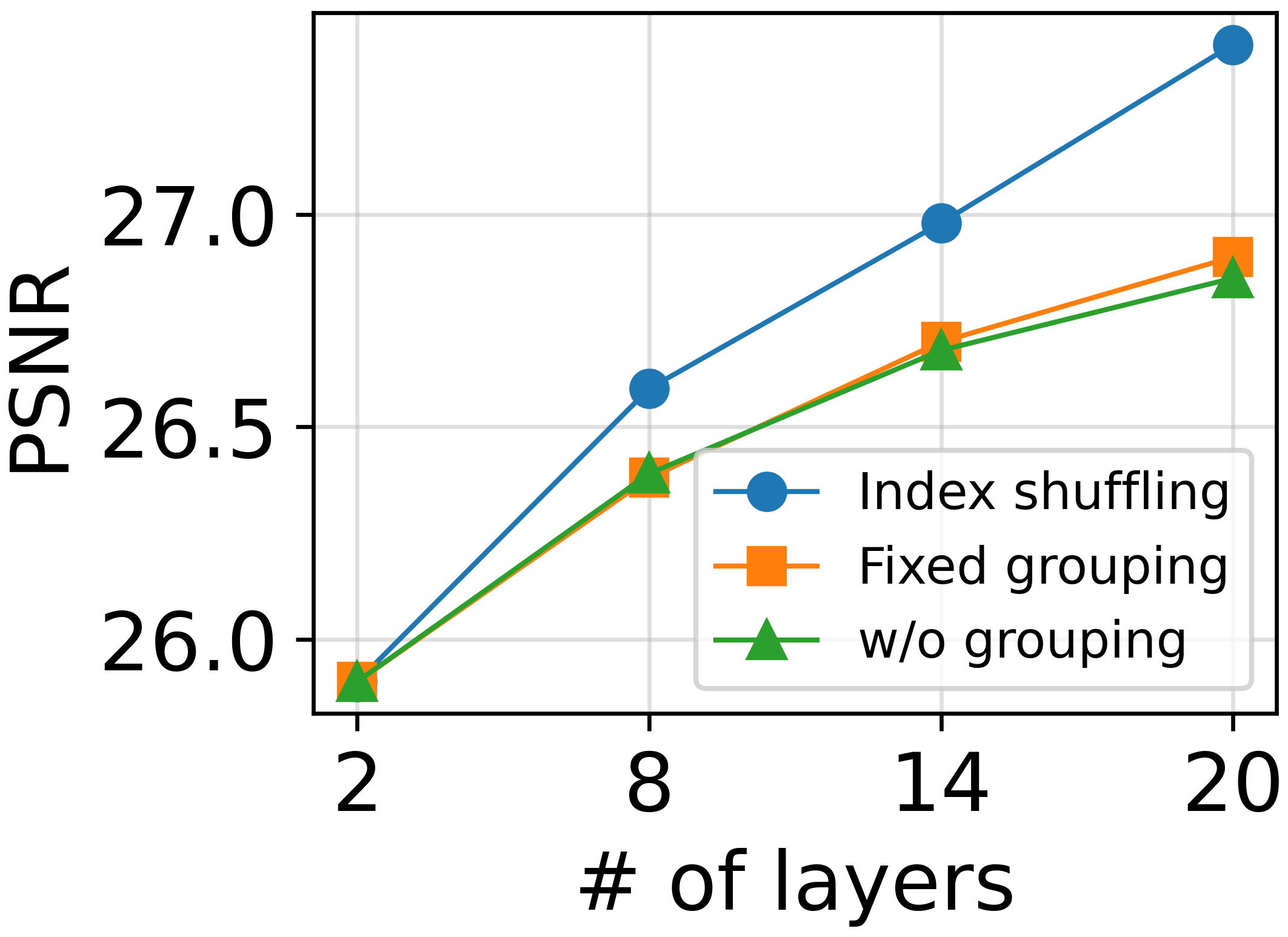}
    \caption{Grouping ablation with varying numbers of logic layers.}
    \label{fig:group}
  \end{minipage}
  \hfill
  \begin{minipage}[t]{0.45\linewidth}
    \centering
    \includegraphics[width=.85\linewidth]{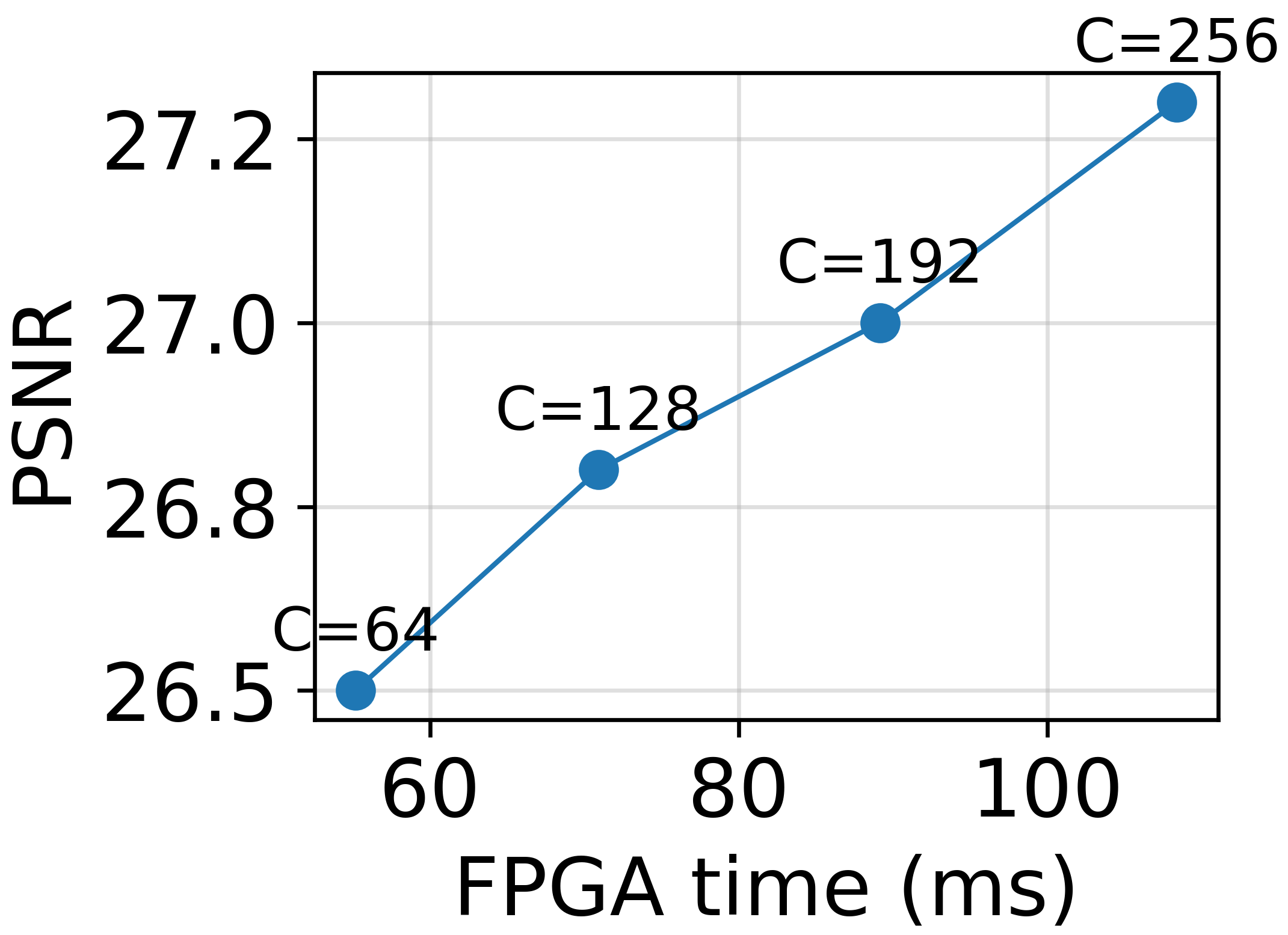}
    \caption{FPGA runtime of LogicIR variants with different channel widths.}
    \label{fig:fpga}
  \end{minipage}
\end{figure}

\noindent \textbf{Component ablation.}
\cref{tab:ablation} presents an ablation study on the key components of LogicIR-S. Starting from a basic stacked convolutional logic gate network, adding the bit decoding module yields a large PSNR gain by converting binary activations into continuous residuals. Incorporating the UNet backbone and index shuffling further enhances hierarchical and inter-group feature learning, while the MSB loss and STE fine-tuning provide additional improvements. In particular, \cref{fig:group} shows that fixed grouped logic layers quickly saturate because they repeatedly operate within the same channel groups, making the model behave like isolated sub-networks. In contrast, index shuffling redistributes features across groups between layers, enabling richer inter-group communication and consistent improvements as more logic layers are stacked. This is further supported by \cref{fig:ablation}, where index shuffling promotes inter-group information exchange and helps each group learn more meaningful representations. Together, these components contribute complementary benefits, achieving the best overall denoising performance.

\begin{table}[!t]
    \caption{Complexity analysis of the denoising models.}
    \label{tab:complexity}
    \centering
    \scriptsize
  \setlength{\tabcolsep}{6pt}
  \renewcommand{\arraystretch}{.9}
    \begin{tabular}{lrrrrr}
        \toprule
        \multirow{2}{*}{Method} &  \multicolumn{1}{c}{\multirow{2}{*}{BOPs}} & \multicolumn{1}{c}{Energy} & \multicolumn{1}{c}{Area} & \multicolumn{1}{c}{Runtime} & \multicolumn{1}{c}{\multirow{2}{*}{PSNR}} \\
        &&\multicolumn{1}{c}{(mJ)}&\multicolumn{1}{c}{(mm$^2$)}&\multicolumn{1}{c}{(ms)}\\
        \midrule
        BBCU-lite~\cite{bbcu}        & 1 097.2 G & 0.69& 2.44& 571.6 & 27.62 \\ 
        HKLUT~\cite{hklut}      &  499.4 G & 0.31& 1.11& 71.2 & 27.34 \\ 
        \rowcolor{teal!10} LogicIR-S & \bf{41.4 G} & \bf{0.03} & \bf{0.09} & \bf{28.2} & 27.40  \\ 
        \rowcolor{teal!10} LogicIR-S-4RT & 169.3 G & 0.11& 0.38& 112.6 & \bf{27.71}  \\ 
        \bottomrule
    \end{tabular}
\end{table}

\noindent \textbf{Color-processing ablation.}
\cref{tab:ablation_color} compares three color-processing strategies. Y-only restores only the luminance channel in the YCbCr domain, yielding the lowest cost (35.9G BOPs) but the lowest quality at 22.63 dB. RGB-joint widens the network from $C$ to $3C$ to process RGB jointly, improving quality to 22.79 dB at a much higher cost of 169.1G BOPs. RGB-channel instead processes channels separately with shared parameters across R/G/B, retaining most of the gain at 22.75 dB with far fewer BOPs (92.7G). We therefore use RGB-channel by default for the best performance--efficiency trade-off.

\subsection{Complexity and Hardware Deployment}
\noindent \textbf{Complexity analysis.} 
To further evaluate the practical efficiency of LogicIR, we analyze not only the operation count but also latency, energy, and hardware area. As shown in~\cref{tab:complexity}, LogicIR-S achieves a runtime of 28.2 ms, which is 20.3$\times$ faster than BBCU-lite (571.6 ms) and 2.5$\times$ faster than HKLUT (71.2 ms). Even with the rotational ensemble (LogicIR-S-4RT), the runtime remains competitive at 112.6 ms, achieving higher PSNR than the BNN-based baseline. The runtime is measured on an RTX 4090 GPU using the Difflogic library~\cite{difflogic}. Following the methodology of~\cite{mulut,neseem2024pikelpn}, LogicIR-S consumes only 0.03 mJ, compared with 0.69 mJ for BBCU-lite and 0.31 mJ for HKLUT, demonstrating its superior energy efficiency. In terms of chip area based on the TSMC N5 process technology, LogicIR-S requires only 0.09 mm$^2$, far smaller than HKLUT (1.11 mm$^2$) and BBCU-lite (2.44 mm$^2$), supporting its practicality for edge deployment.

\noindent \textbf{Practical FPGA deployment.}
We implement LogicIR variants of different sizes on real FPGA devices, following the FPGA implementation protocol of DiffLogic. Since LogicIR is scalable along the channel dimension, we vary the main UNet backbone width $C$ from 64 to 256, while keeping the final layer before bit counting fixed at 2048 channels. We verify these variants on an Intel Cyclone V C9 FPGA under the 4RT inference setting, where four rotated inputs are processed for rotational ensembling. As shown in~\cref{fig:fpga}, the resource usage increases gradually as $C$ increases, indicating that channel width provides a simple and effective knob for adapting LogicIR to different hardware budgets.

\section{Conclusion} 

In this work, we introduced LogicIR, the first logic gate network specifically designed for image restoration. Unlike conventional neural networks that rely on arithmetic operations and floating-point representations, LogicIR operates with discrete logic gates, enabling highly efficient inference. Our architecture incorporates several key components, including a UNet backbone composed of convolutional logic layers, a differentiable bit decoding module, index shuffling for inter-group communication, and MSB-plane-guided supervision. Through extensive experiments on image denoising, deblocking, and deraining tasks, LogicIR demonstrated superior performance compared to existing lightweight approaches such as BNNs and LUT-based baselines, while significantly reducing the required number of operations. These results highlight the potential of LogicIR for edge and resource-constrained deployment under tight resource budgets.

\noindent \textbf{Limitations.} While LogicIR demonstrates strong performance across several image restoration benchmarks, it also presents certain limitations that require further investigation. This study focuses solely on restoration tasks in which the input and output share the same spatial resolution. However, tasks such as image super-resolution involve explicit upsampling, which poses additional architectural and algorithmic challenges. Designing high-quality and scalable upsampling mechanisms within a logic gate framework remains an open and important direction for future research.

\section*{Acknowledgements}
This work was supported by the National Research Foundation of Korea (NRF) grant funded by the Korea government (MSIT) (No. RS-2025-16068196), and Samsung Electronics Company Ltd., under Grant IO230313-05434-01.

%
%
\bibliographystyle{splncs04}
\bibliography{main}
\end{document}


\title{LogicIR: Logic Gate Networks for Image Restoration\texorpdfstring{\\[0.25em]
{\large Supplementary Material}}{ - Supplementary Material}}

\titlerunning{LogicIR}

\author{Hongjae Lee\orcidlink{0000-0001-5312-139X} \and
Myungjun Son\orcidlink{0009-0005-4577-4821} \and
Jaeseong Yu\orcidlink{0009-0005-1427-6651} \and
Seung-Won Jung\thanks{Corresponding author.}\orcidlink{0000-0002-0319-4467}}

\authorrunning{H. Lee \etal}

\institute{Korea University\\
\email{\{jimmy9704, sonbill, jsyu624, swjung83\}@korea.ac.kr}}

\appendix

\maketitle

We organize our supplementary material as follows:
\begin{itemize}
    \item \cref{sec:llie} presents results on low-light image enhancement (LLIE).
    \item \cref{sec:metrics} provides evaluations using various image quality metrics.
    \item \cref{sec:activation} analyzes the layer-wise binary activation distributions.
    \item \cref{sec:msb_planes} presents different configurations of the MSB loss, including its weight $\lambda$ and the number of bit planes used.
    \item \cref{sec:gatecount} discusses how BOPs are estimated for BNNs, LUTs, and LGNs.
    \item \cref{sec:qualitative} provides additional qualitative results for image denoising, deblocking, and deraining.
\end{itemize}

\section{Evaluation on Low-light Image Enhancement}
\label{sec:llie}

\begin{table}[!h]
  \caption{Quantitative comparison on the LOL~\cite{Chen2018Retinex} dataset for low-light enhancement.}
  \label{tab:lol}
  \centering
  \setlength{\tabcolsep}{10pt}
  \renewcommand{\arraystretch}{1.}
  \begin{tabular}{l rcc}
    \toprule
    Method & BOPs & PSNR & SSIM \\
    \midrule
    Bi-Real~\cite{bireal}          & 1 893.4 G & 17.96 & 0.6597 \\
    BBCU-lite~\cite{bbcu}        & 1 899.0 G & 18.42 & 0.6912 \\
    ReActNet~\cite{liu2020reactnet}         & 2 269.5 G & 18.20 & 0.7011 \\
    \midrule
    HKLUT~\cite{hklut}            & 1 227.0 G & 18.08 & 0.7040 \\
    \midrule
    \rowcolor{teal!10} LogicIR-S      & \bf{142.1 G} & 18.22 & 0.6985 \\
    \rowcolor{teal!10} LogicIR-S-4RT  & 579.2 G & \bf{18.30} & \bf{0.7326}  \\
    \bottomrule
  \end{tabular}
\end{table}

We further evaluate LogicIR on low-light enhancement (LLIE) using the LOL dataset\cite{Chen2018Retinex}, as shown in \cref{tab:lol}. LogicIR-S achieves performance comparable to the LUT-based method HKLUT, obtaining 18.22 dB PSNR and 0.6985 SSIM, versus 18.08 dB and 0.7040 for HKLUT, while requiring only 11.6\% of its BOPs. With rotational ensembling, LogicIR-S-4RT further improves the results to 18.30 dB PSNR and 0.7326 SSIM. Although its PSNR is slightly lower than that of BBCU-lite (18.42 dB), it achieves the best SSIM among all compared methods while using only 25.5\%--30.6\% of the BOPs required by the BNN-based baselines.

\section{Evaluation with Various Image Quality Metrics}
\label{sec:metrics}

While the main paper summarizes the comparison across image quality metrics, \cref{tab:quality} provides the detailed numerical results for each metric.

\begin{table*}[!h]
  \caption{Comparison of image quality metrics for various methods at a noise level $\sigma = 25$.}
  \label{tab:quality}
  \centering
  \scriptsize
  \setlength{\tabcolsep}{2pt}
  \renewcommand{\arraystretch}{1.2}
  \begin{tabular}{l l rcccccc}
    \toprule
    Dataset & Method & BOPs & PSNR$_\uparrow$ & SSIM$_\uparrow$ & MS-SSIM$_\uparrow$ & LPIPS$_\downarrow$& NIQE$_\downarrow$ & CLIP-IQA$_\uparrow$ \\
    \midrule
    \multirow{6}{*}{BSD68}&BBCU-lite~\cite{bbcu}                  & 1 097.2 G & 27.62 & 0.7489 & 0.9274 & 0.2006 & 21.21 & 0.3537 \\
    &ReActNet~\cite{liu2020reactnet}         & 1 471.4 G & 27.32 & 0.7188 & 0.9206 & 0.2085 & 20.43 & 0.3043 \\
    &HKLUT~\cite{hklut}                      & 499.4 G   & 27.34 & 0.7291 & 0.9169 & 0.2051 & 21.65 & 0.4911 \\
    \noalign{\vskip 2pt}
    \cline{2-9}
    \noalign{\vskip 2pt}
    \rowcolor{teal!10}\cellcolor{white} & LogicIR-S           & \bf{41.4 G}  & 27.40 & 0.7310 & 0.9239 & 0.2121 & \bf{19.91} & 0.5727 \\
    \rowcolor{teal!10}\cellcolor{white}& LogicIR-S-4RT       & 169.3 G   & \bf{27.71} & \textbf{0.7565} & \textbf{0.9285} & \bf{0.1962} & 20.52 & \bf{0.5945} \\
    \midrule
    \midrule
    \multirow{6}{*}{Set12}&BBCU-lite~\cite{bbcu}                  & 1 097.2 G & \bf{28.22} & \bf{0.7787} & \bf{0.9411} & 0.1830 & 21.70 & 0.4132 \\
    &ReActNet~\cite{liu2020reactnet}         & 1 471.4 G & 27.91 &0.7467 & 0.9338 & 0.1922 & 21.95 & 0.3291\\
    &HKLUT~\cite{hklut}                       & 499.4 G   & 27.94 & 0.7448 & 0.9259 & 0.1974 & 21.64 & 0.5572 \\
    \noalign{\vskip 2pt}
    \cline{2-9}
    \noalign{\vskip 2pt}
    \rowcolor{teal!10}\cellcolor{white}& LogicIR-S           & \bf{41.4 G}  & 27.83 & 0.7451 & 0.9340 & 0.1953 & \bf{20.90} & 0.5980\\
    \rowcolor{teal!10}\cellcolor{white}& LogicIR-S-4RT       & 169.3 G   & \bf{28.22} & 0.7742 & 0.9390 &\bf{0.1810} & 21.19 & \bf{0.5993} \\
    \midrule
    \midrule
    \multirow{6}{*}{Urban100}&BBCU-lite~\cite{bbcu} & 1 097.2 G & 26.84 & \bf{0.7828} & \bf{0.9405} & 0.1987 & 19.46 & 0.3632 \\
    &ReActNet~\cite{liu2020reactnet}  & 1 471.4 G & 26.53 & 0.7489 & 0.9325 & 0.2027 & 18.97 & 0.3532 \\
    &HKLUT~\cite{hklut}   & 499.4 G & 26.30 & 0.7396 & 0.9245 & 0.2092 & 19.49 & 0.5009 \\
    \noalign{\vskip 2pt}
    \cline{2-9}
    \noalign{\vskip 2pt}
    \rowcolor{teal!10}\cellcolor{white}&LogicIR-S & \bf{41.4 G} & 26.57 & 0.7521 & 0.9339 & 0.2037 & \bf{18.49} & \bf{0.5715} \\
    \rowcolor{teal!10}\cellcolor{white}&LogicIR-S-4RT & 169.3 G & \bf{26.85} & 0.7754 & 0.9382 & \bf{0.1982} & 18.78 & \bf{0.5715} \\
    \bottomrule
 \end{tabular}
\end{table*}

\section{Analysis of Binary Activation Distribution in LogicIR}
\label{sec:activation}

\begin{figure}[!h]
    \centering
    \includegraphics[width=.8\linewidth]{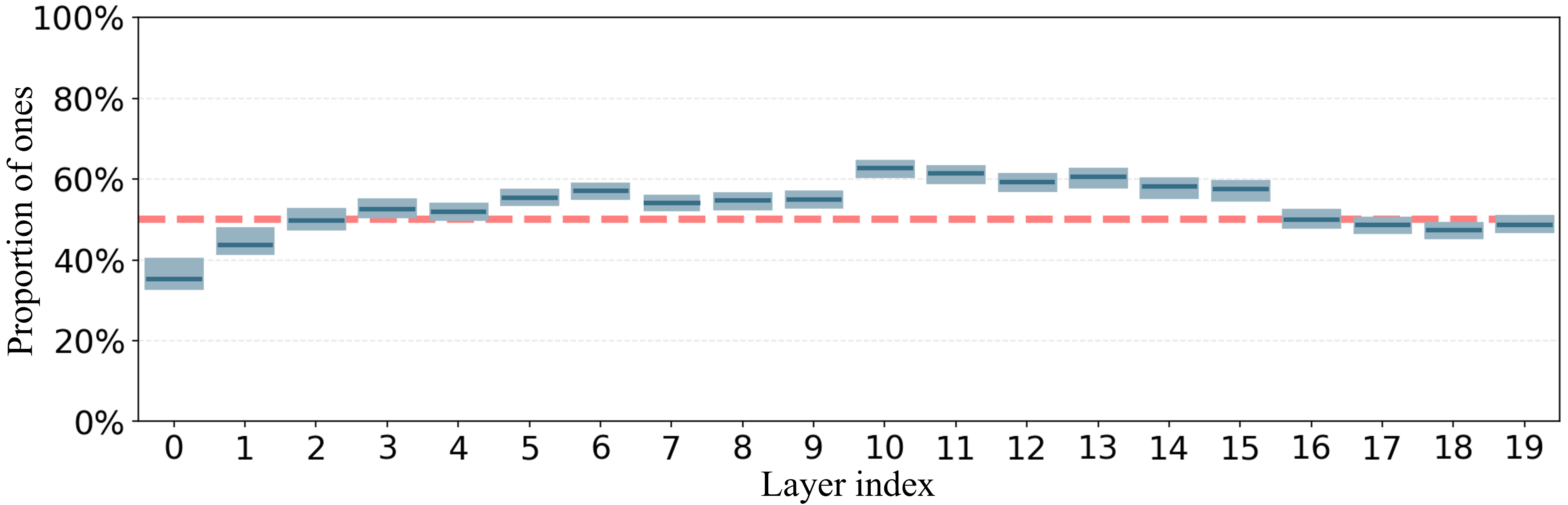}
    \caption{Layer-wise proportion of ones in binary activations on the Set12 dataset.}
    \label{fig:percent}
\end{figure}

To understand the internal mechanisms of LogicIR, we analyze the layer-wise distribution of binary activations. A primary challenge in training logic gate networks (LGNs) is activation saturation. When binary activation values are excessively skewed towards 0 or 1, the entropy of intermediate representations decreases, thereby weakening the gradient flow. To prevent this, differentiable LGNs tend to form a relatively balanced activation distribution during training, where the proportion of 1s remains around 50\%~\cite{convdiff,difflogic}. However, this balance conflicts with image restoration (IR) tasks. While IR requires output distributions that flexibly adapt to spatial content, a 50\% activation ratio centralizes the network's representation, restricting fine-grained pixel predictions. To bridge this gap between training stability and representation flexibility, we propose a learnable scaling-based bit decoding.

Through this architectural design, LogicIR effectively secures the flexible representation capability necessary for IR tasks while preserving the stable training dynamics characteristic of LGNs. Indeed, as illustrated by the layer-wise binary activation distribution in \cref{fig:percent}, the network stably maintains the proportion of 1s near 0.5 across multiple layers, even while successfully performing the image restoration task.

\section{MSB Loss Configurations}
\label{sec:msb_planes}

\begin{figure}[!ht]
  \centering
  \includegraphics[width=.85\linewidth]{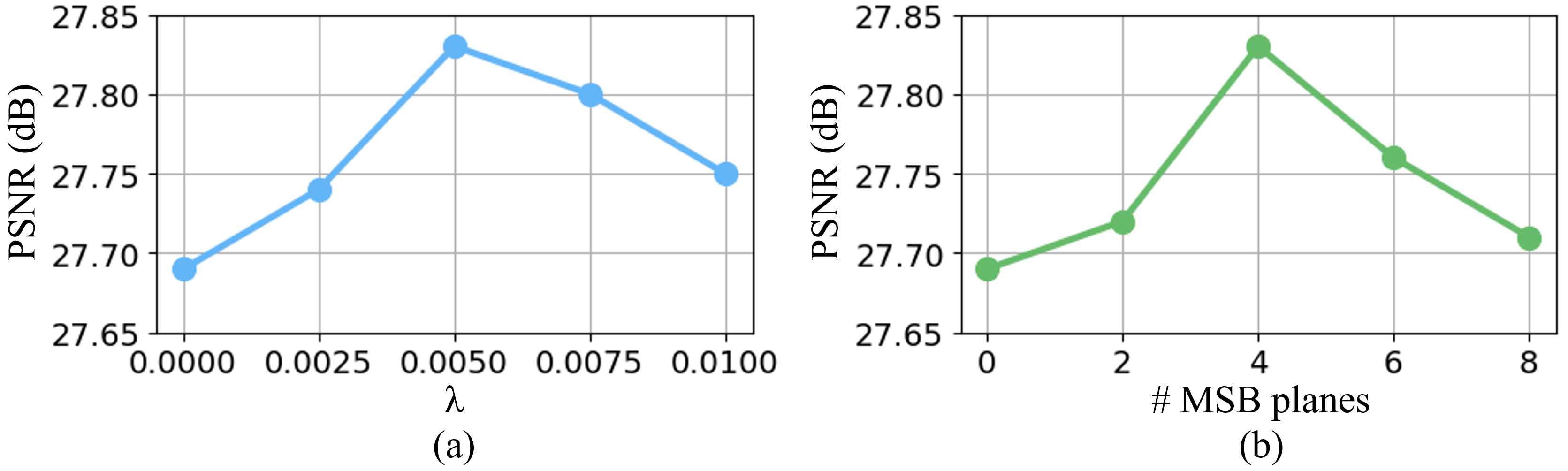}
  \caption{Ablation studies on MSB loss. (a) Effect of MSB loss weight $\lambda$ on denoising performance. (b) Effect of the number of MSB planes used for supervision.}
  \label{fig:msb_ablation}
\end{figure}

We investigate how the design of the MSB (Most Significant Bit) loss influences the performance of LogicIR. Specifically, we analyze the impact of the MSB loss weight $\lambda$ and the number of supervised bit planes, as shown in \cref{fig:msb_ablation}(a) and (b), respectively.

\textbf{Effect of $\lambda$ in Loss Function.}
To assess the contribution of the MSB loss, we vary the weighting factor $\lambda \in \{0, 0.0025, 0.005, 0.0075, 0.01\}$ in Eq.~(7). When $\lambda = 0$, only the standard L2 loss is used, serving as a baseline. As shown in \cref{fig:msb_ablation}(a), increasing $\lambda$ initially improves performance, with the best result observed at $\lambda = 0.005$. Further increasing $\lambda$ (e.g., to $0.01$) slightly degrades performance, likely due to the MSB loss dominating the primary L2 objective.

\textbf{Number of Bit Planes in MSB Loss.}
We further examine how the number of supervised bit planes affects denoising performance. Letting $B \in \{0, 2, 4, 6, 8\}$ denote the number of most significant bits used to construct the MSB reference image, we conduct an ablation study on $B$. As shown in \cref{fig:msb_ablation}(b), using 4 MSB planes achieves the highest PSNR. Using fewer than 4 bits reduces the supervision signal, while using more than 4 introduces noise from low-significance bits, which hinders effective learning for pixel-value regression. Notably, $B = 0$ corresponds to disabling the MSB loss entirely.

\section{Binary Operations in Lightweight Models}
\label{sec:gatecount}

To compare the computational efficiency of various lightweight methods, we report the number of binary operations (BOPs) required for inference. In this section, we describe how the BOPs of BNNs, LUTs, and LGNs are estimated.

\textbf{Binary Operations in BNNs.}
Following previous logic gate network studies~\cite{difflogic,ttnet,convdiff}, we estimate the BOPs of BNNs by decomposing each binary multiply-accumulate (BMAC) operation into an XNOR stage followed by a bitcount operation. Specifically, for an input of $n$ binary values, the multiplication stage requires $n$ XNOR operations, and the bitcount operation requires an additional $7n$ binary operations, resulting in a total of $8n$ BOPs per BMAC operation. This estimate does not include additional operations required for binary thresholding.

\textbf{Binary Operations in LUTs.}
For LUT-based methods, following the approach in~\cite{ttnet}, we convert each learned LUT into a functionally equivalent Boolean expression in conjunctive normal form (CNF), and count the number of binary operations required to evaluate that expression. Since LUT outputs are typically 8-bit, each output bit plane is treated as an independent boolean function. Accordingly, we generate eight separate CNF representations, one for each output bit. We use the Espresso logic minimization algorithm~\cite{espresso} to derive compact CNF formulas. The exact number of BOPs depends on the complexity of each CNF, which is influenced by the structure of the LUT mapping. Simple or repetitive mappings lead to fewer binary operations, whereas more complex mappings require more operations. As a result, the overall BOP count may vary across tasks depending on the input-output complexity of the LUT.

\textbf{Binary Operations in LGNs.}
In LGNs, the BOP count is determined by the operations performed in the convolutional logic layers and the bit decoding module. After training, the actual number of BOPs may vary depending on the specific gate types selected in the logic tree kernels. In particular, some gates may become trivial operations such as constant True or False, effectively eliminating the need for upstream computations. Likewise, buffer gates acting as identity functions allow one branch of the logic tree to be skipped. These structural simplifications reduce the number of effective binary operations required during inference. Our final BOP count reflects these optimizations by counting only the non-trivial binary operations executed at inference time.

\section{Additional Qualitative Results}
\label{sec:qualitative}
\begin{figure*}[!ht]
  \centering
  \includegraphics[width=.98\linewidth]{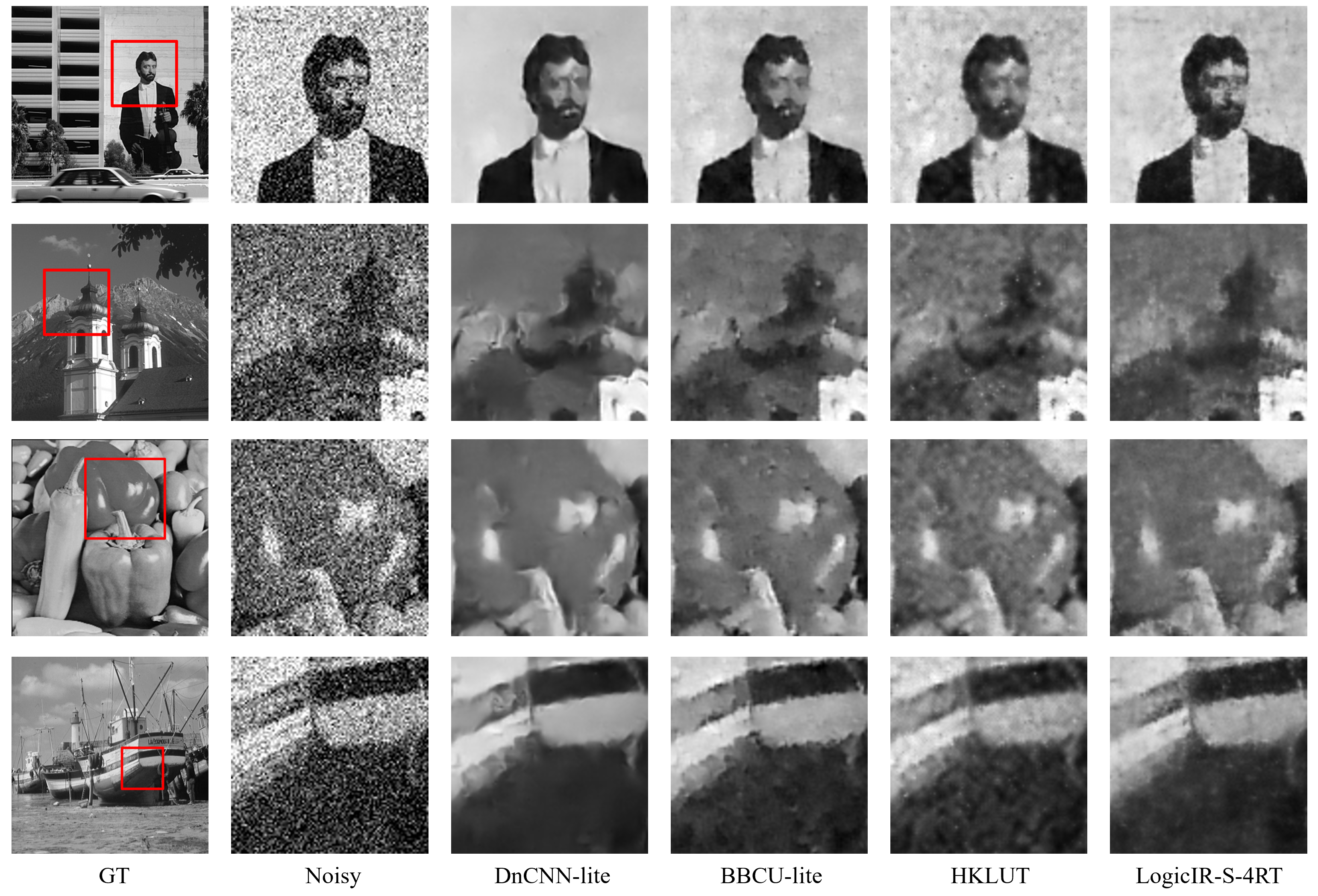}
  \caption{Qualitative results on image denoising with a noise level of $\sigma = 50$.}
  \label{fig:std50}
\end{figure*}

\begin{figure*}[!ht]
  \centering
  \includegraphics[width=.98\linewidth]{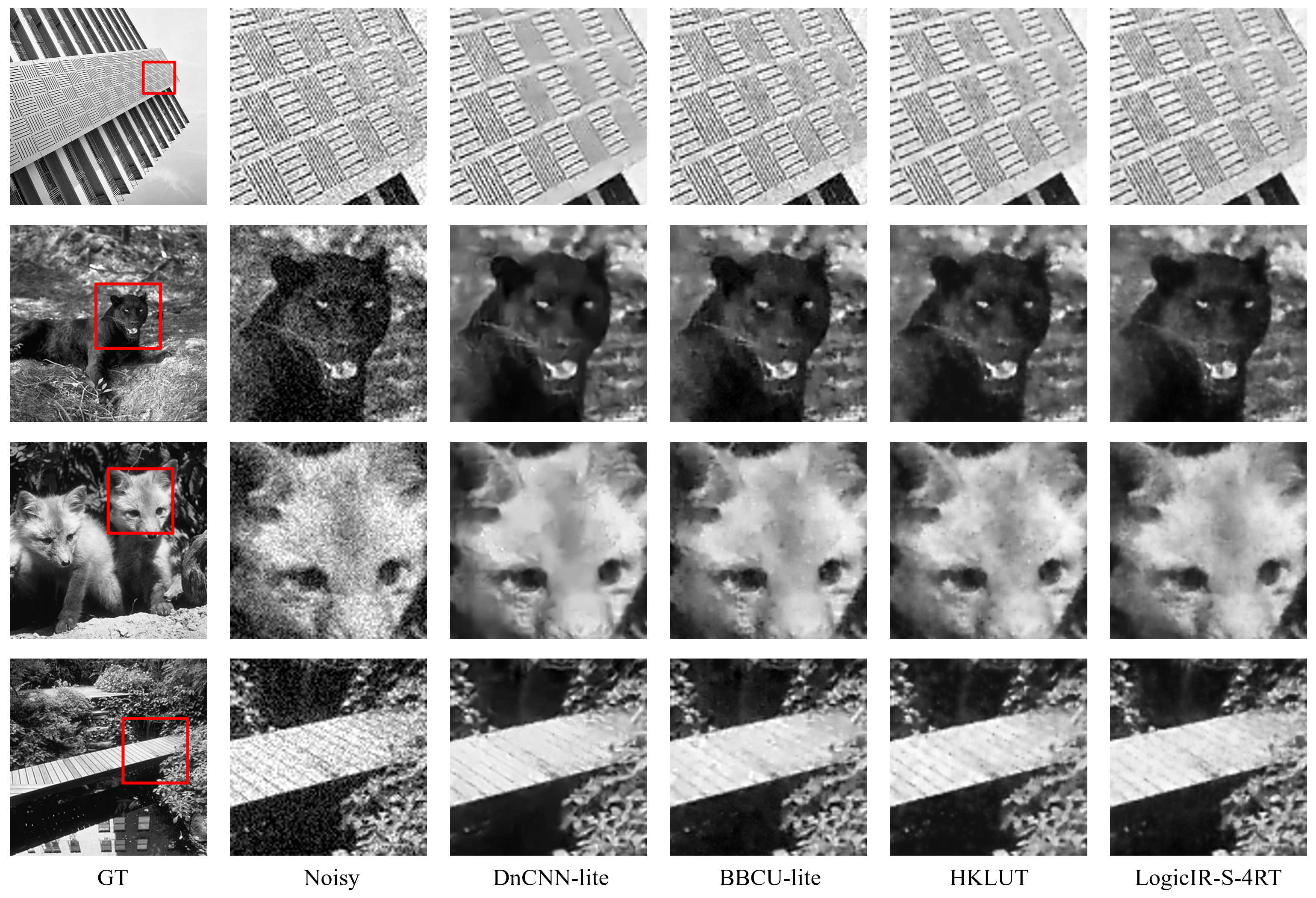}
  \caption{Qualitative results on image denoising with a noise level of $\sigma = 25$.}
  \label{fig:std25}
\end{figure*}

\begin{figure*}[!t]
  \centering
  \includegraphics[width=.98\linewidth]{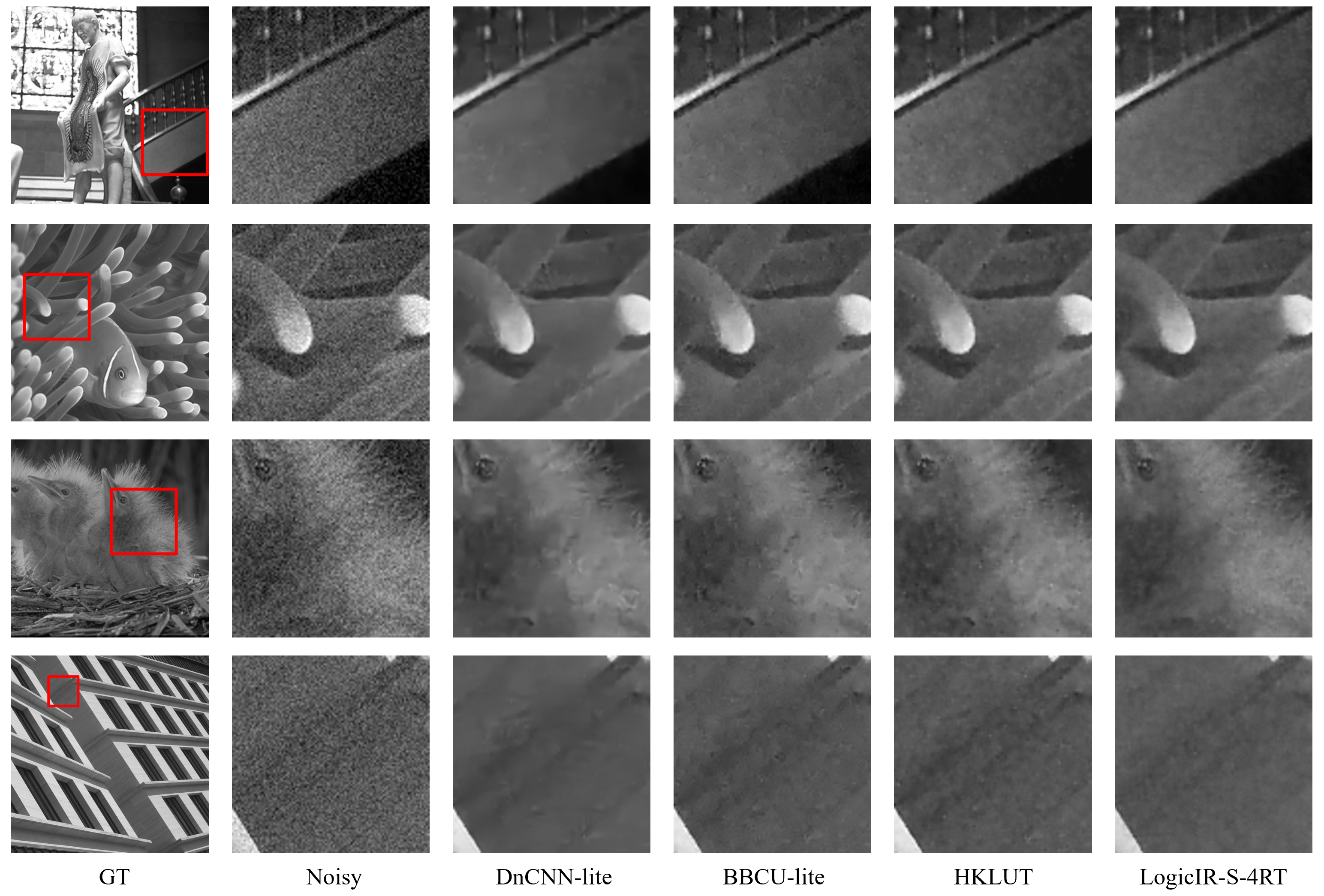}
  \caption{Qualitative results on image denoising with a noise level of $\sigma = 15$.}
  \label{fig:std15}
\end{figure*}

\begin{figure*}[!ht]
  \centering
  \includegraphics[width=.98\linewidth]{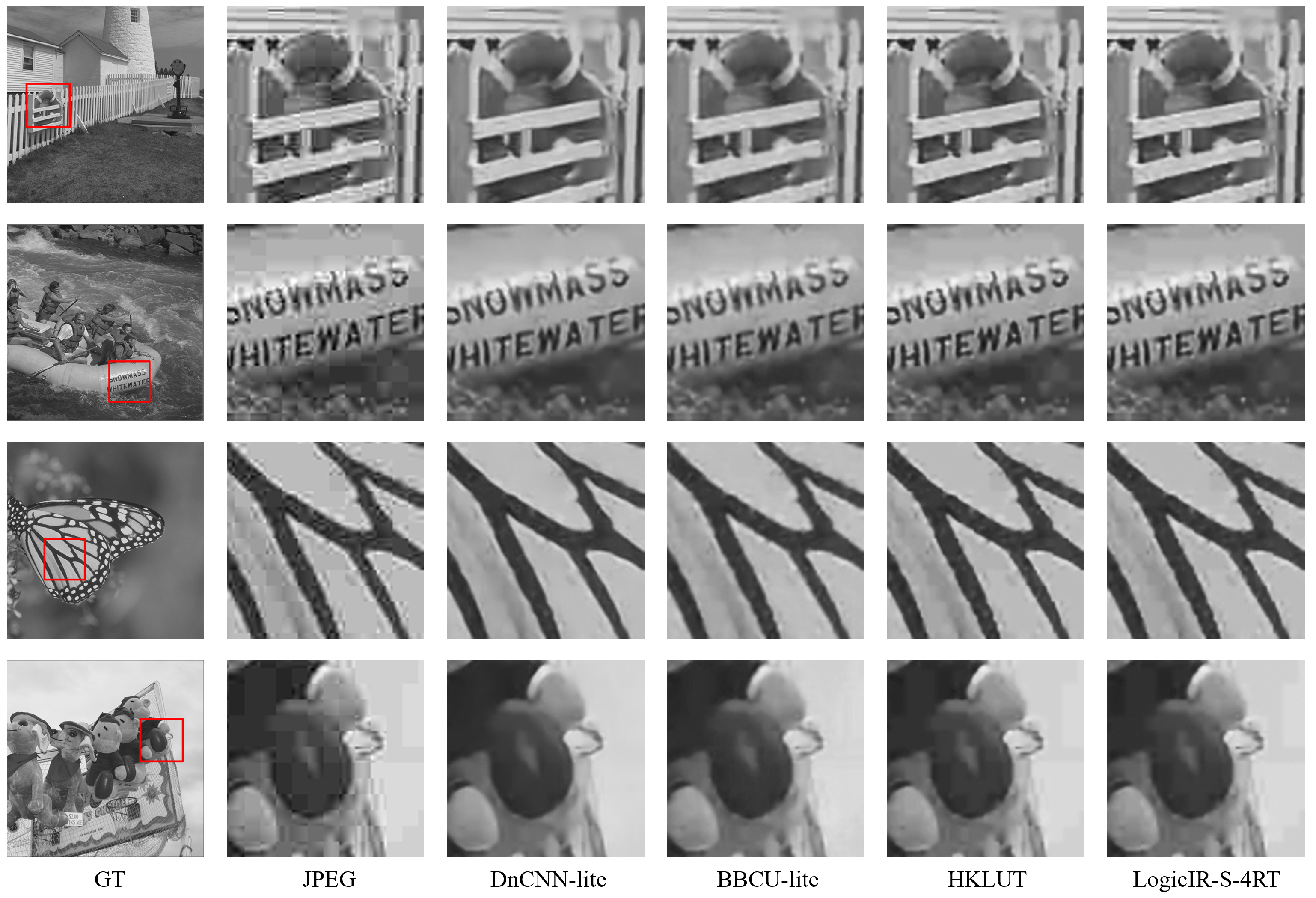}
  \caption{Qualitative results on JPEG deblocking.}
  \label{fig:deblock}
\end{figure*}

\begin{figure*}[!ht]
  \centering
  \includegraphics[width=.98\linewidth]{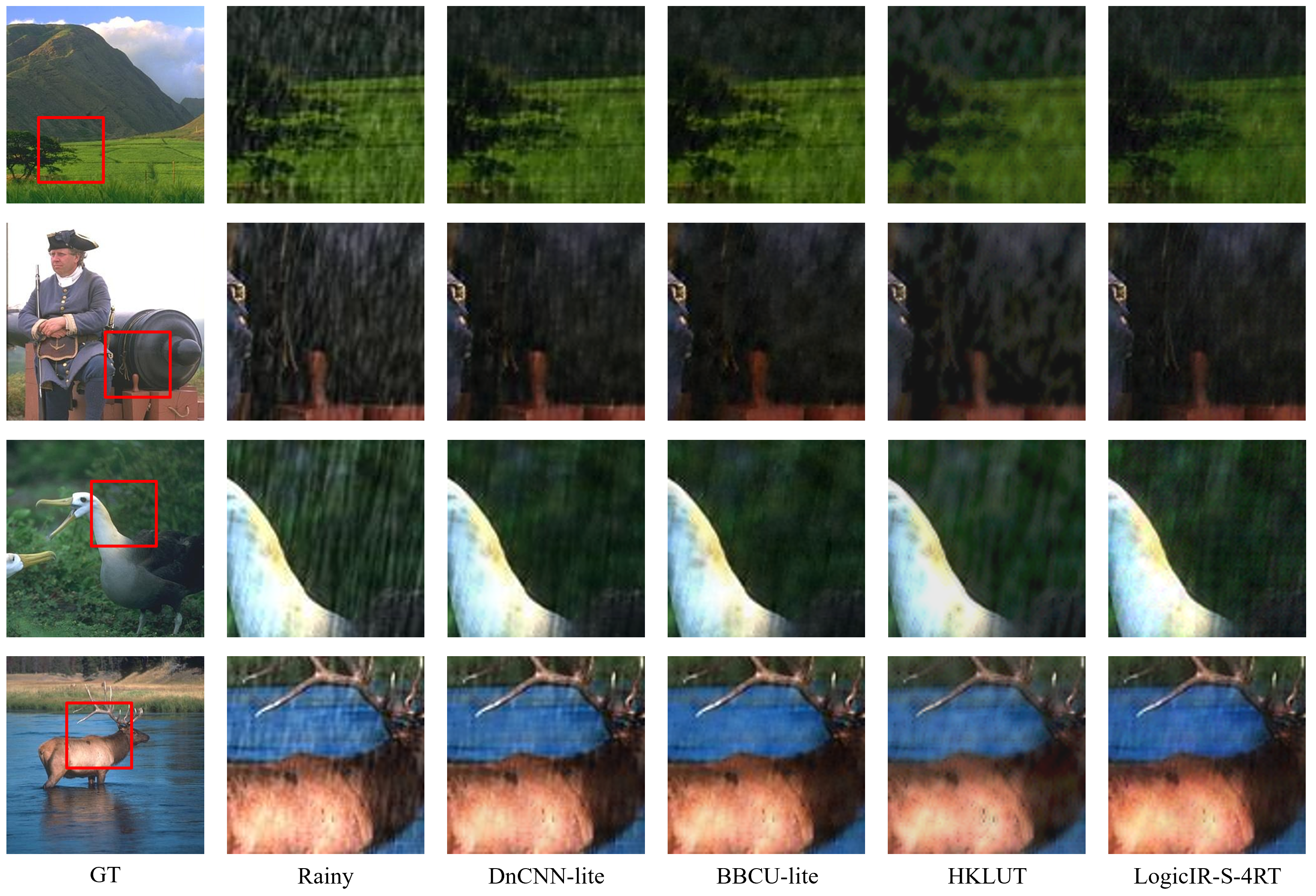}
  \caption{Qualitative results on image deraining.}
  \label{fig:derain}
\end{figure*}

We provide additional qualitative results in~\cref{fig:std15,fig:std25,fig:std50,fig:deblock,fig:derain} to further demonstrate the effectiveness of LogicIR across various image restoration tasks.

%
%
\bibliographystyle{splncs04}
\bibliography{main}